\documentclass{article}

\PassOptionsToPackage{numbers,sort,square,comma}{natbib}
\usepackage[preprint]{neurips_2021}

\usepackage{microtype}
\usepackage{graphicx}
\usepackage{subfigure}
\usepackage{amsfonts}
\usepackage{amsmath}
\usepackage{mathtools}

\usepackage{bm}
\usepackage{multirow}
\usepackage{enumitem}
\usepackage[font=small,labelfont=bf,tableposition=top]{caption}

\usepackage{booktabs} 
\usepackage{xcolor}
\usepackage{bbm}
\usepackage{sidecap}

\usepackage{lmodern}
\usepackage{slantsc}

\usepackage{graphicx}

\def\fauxschelper#1 #2\relax{%
  \fauxschelphelp#1\relax\relax%
  \if\relax#2\relax\else\ \fauxschelper#2\relax\fi%
}
\def\Hscale{.85}\def\Vscale{.74}\def\Cscale{1.12}
\def\fauxschelphelp#1#2\relax{%
  \ifnum`#1>``\ifnum`#1<`\{\scalebox{\Hscale}[\Vscale]{\uppercase{#1}}\else%
    \scalebox{\Cscale}[1]{#1}\fi\else\scalebox{\Cscale}[1]{#1}\fi%
  \ifx\relax#2\relax\else\fauxschelphelp#2\relax\fi}

\definecolor{orange}{rgb}{1,0.5,0}
\definecolor{light_red}{rgb}{0.94,0.5,0.5}
\definecolor{blue}{rgb}{0,0,1.0}
\definecolor{orange}{RGB}{255, 133, 51}

\definecolor{CascadedTDColor}{RGB}{182,54,121}
\definecolor{CascadedCEColor}{RGB}{110,30,121}
\definecolor{SerialTDColor}{RGB}{77,184,255}
\definecolor{SerialTDZColor}{RGB}{127,255,0} 
\definecolor{SerialCEColor}{RGB}{54,129,179}

\definecolor{bestGreen}{RGB}{0, 200, 0}
\definecolor{MetaCogColor}{RGB}{128,128,0}

\definecolor{SequentialColor}{RGB}{201,109,77}
\newcommand{\colorSequential}[1]{\textcolor{SequentialColor}{#1}}

\definecolor{highlightColor}{RGB}{51,133,255}
\newcommand{\highlight}[1]{\colorbox{highlightColor!20}{#1}}

\newcommand{\colorCascadedTD}[1]{\textcolor{CascadedTDColor}{#1}}
\newcommand{\colorCascadedCE}[1]{\textcolor{CascadedCEColor}{#1}}
\newcommand{\colorSerialTDZ}[1]{\textcolor{SerialTDZColor}{#1}}
\newcommand{\colorSerialTD}[1]{\textcolor{SerialTDColor}{#1}}
\newcommand{\colorSerialCE}[1]{\textcolor{SerialCEColor}{#1}}

\newcommand{\CascadedTD}{{\footnotesize \sc \colorCascadedTD{CascadedTD}}}
\newcommand{\CascadedCE}{{\footnotesize \sc \colorCascadedCE{CascadedCE}}}

\newcommand{\SerialTD}{{\footnotesize \sc \colorSerialTD{SerialTD}}}
\newcommand{\SerialCE}{{\footnotesize \sc \colorSerialCE{SerialCE}}}
\newcommand{\SerialTDMH}{{\footnotesize \sc \colorSerialTD{SerialTD-MultiHead}}}
\newcommand{\MultiHead}{{\footnotesize \sc {MultiHead}}}
\newcommand{\SingleHead}{{\footnotesize \sc {SingleHead}}}

\newcommand{\Sequential}{{\small \sc \colorSequential{SerialCE$\times9$}}} 
\newcommand{\MetaCog}{{\small \sc {\textcolor{MetaCogColor}{MetaCog}}}}

\usepackage{hyperref}

\title{Improving Anytime Prediction \\with Parallel Cascaded Networks\\and a Temporal-Difference Loss}

\author{%
  Michael L. Iuzzolino\\
  Google Research, Brain Team and\\
  Department of Computer Science, University of Colorado\\
  \AND
  Michael C. Mozer\\
  Google Research, Brain Team\\
  \AND
  Samy Bengio\\
  Google Research, Brain Team\thanks{Currently at Apple.}\\
}

\begin{document}

\maketitle

\begin{abstract}
Although deep feedforward neural networks share some characteristics with the primate visual system, a key distinction is their dynamics.  Deep nets typically operate in \emph{serial} stages wherein each layer completes its computation before processing begins in subsequent layers.   In contrast, biological systems have \emph{cascaded} dynamics: information propagates from neurons at all layers in parallel but transmission occurs gradually over time, leading to speed-accuracy trade offs even in feedforward architectures. We explore the consequences of biologically inspired parallel hardware by constructing cascaded ResNets in which each residual block has propagation delays but all blocks update in parallel in a stateful manner. Because information transmitted through skip connections avoids delays, the functional depth of the architecture increases over time, yielding  anytime predictions that improve with internal-processing time. We introduce a temporal-difference training loss that achieves a strictly superior speed-accuracy profile over standard losses and enables the cascaded architecture to outperform state-of-the-art anytime-prediction methods. The cascaded architecture has intriguing properties, including: it classifies typical instances more rapidly than atypical instances; it is more robust to both persistent and transient noise than is a conventional ResNet; and its time-varying output trace provides a signal that can be exploited to improve information processing and inference.
\end{abstract}

\label{sec:introduction}
Since the earliest investigations of artificial neural nets, their design has been informed by biological neural 
nets  \cite{McCullochPitts}. Perhaps the most compelling example is the convolutional 
net for machine vision, which has adopted properties of primate cortical neuroanatomy 
including a hierarchical layered organization, local receptive fields, and spatial 
equivariance \cite{Fukushima1980}. 
In this article, we investigate computational consequences of two fundamental properties of biological 
information processing systems that have not been considered in the design of deep neural nets. First,
\emph{the brain consists of massively parallel, dedicated hardware with neurons throughout the cortex updating continuously and simultaneously.} Second, \emph{information transmission between neurons introduces time delays} \cite{Bialek1992}.
As a result, unrefined and possibly incomplete neural state in one region is transmitted
to the next region even as the state evolves; and feedforward connectivity yields
a speed-accuracy trade off in which the initial response to a static input occurs rapidly 
but can be inaccurate,  with the output gradually improving over internal processing time.
Following McClelland \cite{mcclelland1979time}, we refer to such an architecture as \emph{cascaded}.

Cascaded dynamics contrast sharply with the dynamics of standard feedforward nets, which operate
in \emph{serial} stages, each layer completing its computation before subsequent layers begin processing.
Cascaded dynamics are also quite different than the dynamics of vision models with recurrent
connections \cite[e.g.,][]{Kar2019,Kriegeskorte2015,mcintosh2018recurrent,Spoerer2017}, which, given
a static input, may iteratively update, but layer updates are still computed serially with each 
layer completing its computation and then feeding it immediately to the next layer (or back to itself).
Fundamentally, our investigation asks: Supposing we take a step toward biological realism with massively 
parallel hardware and relatively slow inter-neuron communication, what are the computational benefits and
consequences?\footnote{Like much other research in deep learning \cite{Kriegeskorte2015, dicarlo2012does}, biology informs our work by providing novel forms of inductive bias. Our goal is to investigate computational consequences of these biases, not to model biological phenomena per se.}

\begin{figure}[!t]
    \centering
    \includegraphics[width=0.9\textwidth]{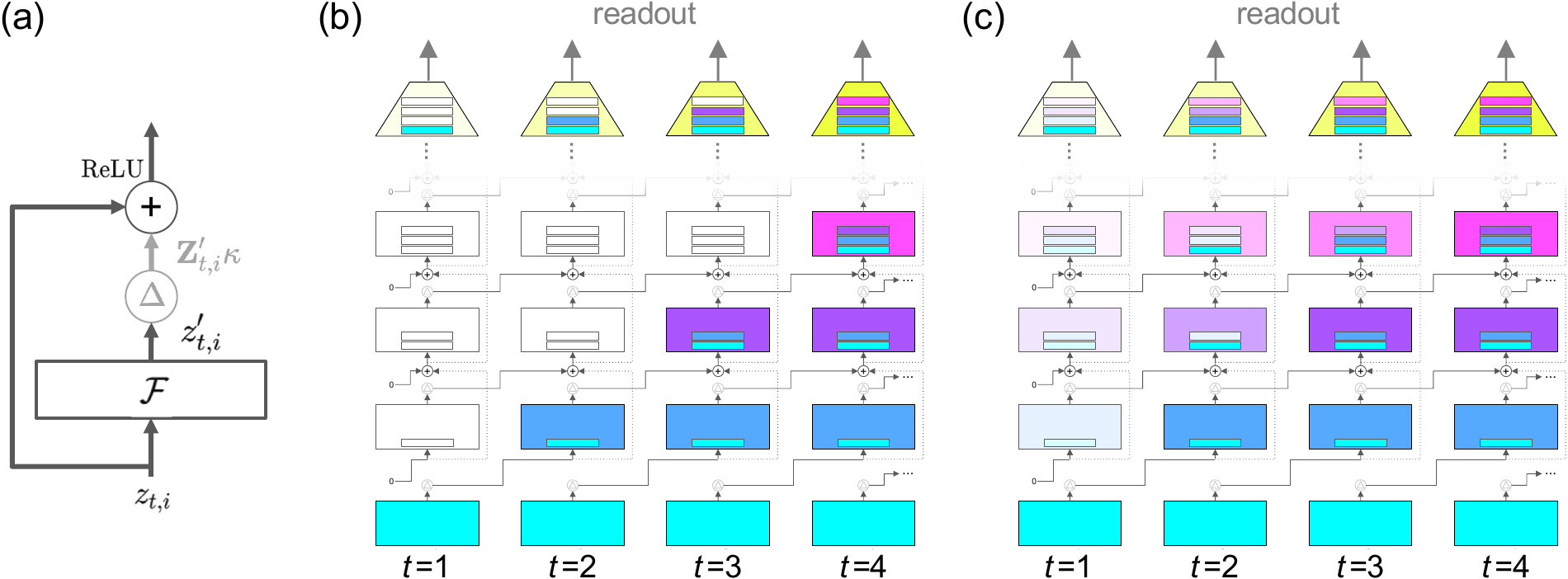}
    \small
    \caption{(a) ResNet building block, with additional delay component ($\Delta$, in grey) that convolves
    a temporal kernel with the block output. Details in text. (b) A standard 
    \emph{serial} ResNet is unrolled in time, with columns depicting time slices.
    Each rectangle is a ResNet block, which may consist of two or more convolutional layers.
    In the serial model, blocks are updated sequentially.  Blocks which have not yet been
    activated are colored white and blocks which have been activated are shown in a hue
    unique to that block. The input is \color{cyan}cyan\color{black}.
    The narrow bars within each block signify the activation state of all blocks below that  
    are contributing to the block's state (via skip connections). Read out from the model is via the yellow trapezoid at the top, which enables anytime prediction. The narrow bars
    inside the trapezoid indicate the block information available at each time for classification (via skip connections).
    (c) A \emph{cascaded} ResNet is unrolled in time. In the cascaded model, all blocks
    update in parallel; however, at each step, they may rely on partial updates of lower 
    blocks.  
    As a result, multiple processing steps are required for a layer's activation to reach its 
    asymptotic state.
    The color intensity (saturation) 
    of a block indicates how close a block's activation state is to its asymptotic state. 
    }
    
    \label{fig:tdl_resnet}
\end{figure}

We construct cascaded networks by introducing propagation delays in deep feedforward nets provided with a static input.
We treat the net as massively parallel such that all units across all layers are updated
simultaneously and iteratively. We focus on the ResNet architecture \cite{he2016deep} and we introduce a propagation 
delay into each residual block (Figure~\ref{fig:tdl_resnet}a). Because the skip connection permits faster transmission 
of more primitive perceptual representations, the functional depth of the resulting architecture increases 
over internal-processing time, yielding a trade off between processing speed and complexity of processing.
Consequently, the architecture offers a natural, integral mechanism for making predictions at any point in processing, known as \emph{anytime prediction}
\cite{zilberstein1996using}. 
Speed-accuracy trade offs are a fundamental 
characteristic of human information processing  \cite{JonesKinoshitaMozer2009,RatcliffMcCoon2008} and human perception 
has been modeled with deep learning anytime prediction methods \cite{Kumbhar2020}.

Although we focus on the ResNet, our approach can be incorporated into any model with skip connections 
(e.g., Highway Nets \cite{srivastava2015highway}, DenseNet \cite{huang2017densely}, U-Net \cite{ronneberger2015u},
Transformers \cite{Vaswanietal2017}). The contrast between a \emph{serial}, one-layer-at-a-time model and a 
\emph{cascaded},  parallel-update model is illustrated in  Figures~\ref{fig:tdl_resnet}b and 
\ref{fig:tdl_resnet}c, respectively. To step through the operation of the cascaded model,
at time 1, only the first residual block has received meaningful input, and the model prediction is therefore based 
only on this block's computation. At time 2, \emph{all} higher residual blocks have received input from block 1, 
and the output is therefore based on \emph{all} blocks' computations, though  blocks 2 and above have deficient input. 
At each subsequent time, all blocks are receiving meaningful input, but it is not until time $t$ that block 
$t$ has reached its asymptotic output because its input does not stabilize until $t-1$.
In essence, the cascaded model behaves like a WideResNet \cite{zagoruyko2016wide} on the first steps and then
becomes a deep ResNet.

\begin{figure}[t!]
    \centering
    \includegraphics[height=1.65in]{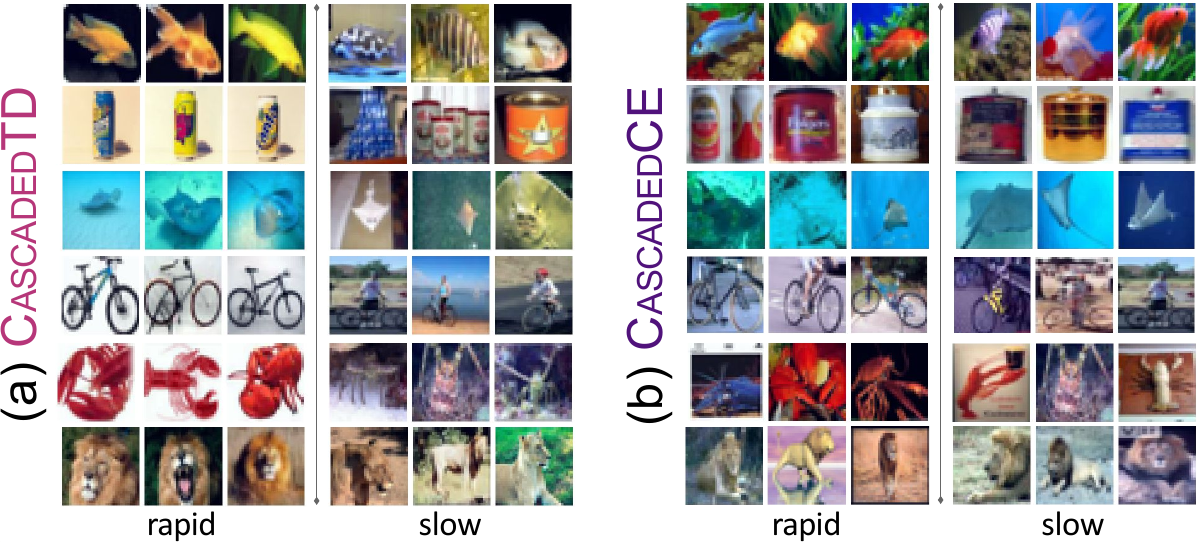}
    \small
    \caption{(a) CIFAR-100 instances categorized rapidly (left) and slowly (right) by a cascaded model trained
    using a TD loss. (b) same as (a) for a standard cross-entropy loss.
    The cascaded model with a TD loss stratifies instances by typicality, with rapid processing of prototypical
    views on a homogenenous background and slow processing of unusual and cluttered views.}
    \label{fig:rapidslow}
\end{figure}

Our work makes the following key contributions.
\vspace{-2mm}
\begin{itemize}[leftmargin=*,itemsep=0pt,parsep=2pt]
\item We demonstrate the superiority of the cascaded architecture to the serial 
(Figures~\ref{fig:tdl_resnet}b,c), indicating that parallelism can be exploited in 
a way that has not previously been studied.
\item We propose and evaluate a novel training objective aimed at improving the predictions of anytime models.
This \emph{temporal-difference (TD) loss} \citep{sutton1988learning} encourages
the most accurate response as quickly as possible. TD training improves the performance of both cascaded 
and serial architectures. Although a rich literature exists aimed at reducing the number of
computational steps required to obtain an accurate answer
\cite{bolukbasi2017adaptive, bulat2020toward, bulat2017far, carreira2018massively, fischer2018streaming, hu2018anytime,  hu2019learning,  huang2018, huang2017multi,  iuzzolino2019convolutional, kaya2019shallow, lee2018anytime, lee2018anytime, mcintosh2018recurrent, newell2016stacked, scardapane2020should, teerapittayanon2016branchynet, yang2020resolution, zamir2017feedback},  
all of this work uses a degenerate form of TD
for training and our results suggest that these models can be improved using TD.
\item The cascaded model trained with TD (\CascadedTD) tends to respond most rapidly to prototypical 
exemplars, whereas training with the standard cross-entropy loss (\CascadedCE) does not (Figure~\ref{fig:rapidslow}). We
assess with three quantitative prototypicality measures, and we further show that \CascadedTD\ rapidly converges on 
the correct semantic family, whereas \CascadedCE\ does not. These facts indicate that \CascadedTD\ organizes
knowledge differently across layers than does \CascadedCE.
\item  We show that \CascadedTD\ obtains a strictly superior speed-accuracy profile compared to previously proposed
anytime prediction models, which are all based on a serial architecture.
\item We demonstrate other virtues of \CascadedTD: it is more robust to input noise, and its time-varying
output trace provides useful signals for \emph{meta-cognitive} processes---separately trained
nets that make judgments about the cascaded architecture's accuracy. 
\end{itemize}


\section*{Related Work}
\label{sec:related_works}

\textit{Prior research on cascaded models.}
From a psychological perspective, McClelland \cite{mcclelland1979time} characterizes human mental computation 
in terms  of a hierarchy of leaky integrators that continually transmit partial information as it becomes available.
We are aware of no work in deep learning on static image processing with cascaded models, but there exist two
investigations focused on video sequence processing, where the model state from the previous frame
is used to efficiently process the next. 
Fischer et al.~\cite{fischer2018streaming} present a \emph{streaming rollout} framework for recurrent nets and they
very briefly explore the temporal dynamics of cascaded models, showing benefits to early predictions. They present a general taxonomy that includes our proposed feedforward cascaded model, but their focus is almost entirely on formal definitions and a framework that lays out the space of all well-formed roll out patterns (update orders). In contrast, our focus is almost entirely on training procedures that leverage the dynamics of cascaded models, on early read-out mechanisms, and on the computational consequences of these training and read-out mechanisms. Kugele et al.~\cite{kugele2020efficient} focus on spiking neural net dynamics, on time-varying inputs, and on reductions in latency that are obtained as a sequence unfolds due to autocorrelations in the input sequence. Notably, Kugele et al. explore a variety of heuristic training losses and they settle on TD(1) as their preferred loss, but they do not explore the rest of the TD($\lambda$) family. Our work is complementary.

Carreira et al.~\cite{carreira2018massively} present a causal video understanding model that performs depth-parallel computation with the objective of 
improving video processing efficiency via the maximizing throughput, minimizing latency, and reducing clock cycles. 

\textit{Recurrent nets for vision}. Recurrent nets have been used in vision
\cite[e.g.,][]{iuzzolino2019convolutional,Kar2019,Kubilius2018,Kriegeskorte2015,leroux2018,Spoerer2017,Spoerer2020}, which adds a dimension of internal
processing time for every external input (see also \cite{Graves2016}). However, these models perform serial layerwise
updates and therefore fundamentally differ in their operation from cascaded models.

\textit{Anytime prediction.}
Anytime prediction models 
\cite{bolukbasi2017adaptive,Elbayad2020,hu2018anytime,hu2019learning,huang2018,huang2017multi,iuzzolino2019convolutional,kaya2019shallow,Larsson2017,lee2018anytime,marquez2018deep,mcintosh2018recurrent,newell2016stacked,scardapane2020should,teerapittayanon2016branchynet,wang2017idk,yang2020resolution,zamir2017feedback}
assume serial operation of layers, but allow for predictions to be made from intermediate layers of the architecture. In the
simplest case, after $t$ steps, $t$ layers have been activated, and at each step, a prediction
is made from the last activated layer \citep[e.g.,][]{hu2019learning,kaya2019shallow}. 
Figure~\ref{fig:tdl_resnet}b illustrates a serial model that performs anytime prediction.
Some of these models have intrinsic stopping criteria \citep[e.g.,][]{Chen2020}; others rely on selection of a
stopping confidence threshold \citep[e.g.,][]{kaya2019shallow,teerapittayanon2016branchynet, yang2020resolution}.





\textit{Temporal-difference learning.} TD learning has a rich history, mostly in the RL community for value function
estimation. TD can be used for supervised learning as well. In fact, the two previous works in deep learning using cascaded models \cite{carreira2018massively, fischer2018streaming} perform a boundary case of supervised TD, TD(1),
which we show to have inferior performance.
A variety of non-cascaded models, both recurrent 
\citep{iuzzolino2019convolutional,  lee2018anytime, mcintosh2018recurrent,  zamir2017feedback}
and feedforward 
\citep{bolukbasi2017adaptive, bulat2020toward, bulat2017far, hu2018anytime,  hu2019learning,  huang2018, huang2017multi,  kaya2019shallow, lee2018anytime, newell2016stacked, scardapane2020should, teerapittayanon2016branchynet,  yang2020resolution}, 
aim to reduce the number of computational steps required to obtain an accurate output. 
All use TD(1) for training. No previous research has explored the general formulation of TD for improving anytime prediction.



\section*{Deep Cascaded Networks}
\label{sec:cascaded_deep_networks}

Many modern deep architectures---including ResNet \citep{he2016deep}, Highway Nets \cite{srivastava2015highway}, DenseNet \cite{huang2017densely}, U-Net \cite{ronneberger2015u}---incorporate skip connections that bypass strictly layered feedforward connectivity, analogous to the architecture of visual cortex \cite{felleman1991distributed}. Under the biological assumption that signals transmitted through a neural layer are delayed relative to signals that bypass the layer, we construct a cascaded model using ResNets by introducing a 
novel computational component that delays the transmission of signals from the output of each computational layer, 
denoted $\Delta$ in Figure~\ref{fig:tdl_resnet}a. Because these delays extend processing in time, the hidden states require a time index. 
The input to ResNet block $i$ at time $t$ is denoted $\bm{z}_{t,i}$. The block transforms this input via the residual transform, yielding
$\bm{z}^\prime_{t,i} = \mathcal{F} ( \bm{z}_{t,i} )$.
We conceive of $\Delta$ as a tapped delay-line memory of the transform history,
$\bm{Z}^{\prime}_{t,i} = [ \bm{z}^{\prime}_{t,i} ~ \bm{z}^{\prime}_{t-1,i} ~\ldots ~\bm{z}^{\prime}_{1,i} ]$,
which is convolved with a temporal kernel $\bm{\kappa}$ to produce the block output
\begin{equation}
\bm{z}_{t,i+1} = \mathrm{ReLU} \left( \bm{z}_{t,i} + \bm{Z}^\prime_{t,i} \bm{\kappa} \right) .
\end{equation}
The kernel $\bm{\kappa} = [1~ 0~ 0 \ldots 0]$ recovers the standard ResNet in which communication between layers is instantaneous.
We consider two kernels to introduce time delays.  With ${\bm{\kappa} = [0~~ 1~~ 0~~ 0~ \ldots 0]}$, a discrete one-step delay is introduced (\emph{OSD} for short).
With $\bm{\kappa} = (1-\alpha)[1~~ \alpha~~ \alpha^2~~ \alpha^3 \ldots ]$, we obtain exponentially weighted smoothing (\emph{EWS} for short), where 
larger $\alpha \in [0,1)$ yield slower transmission times.
Note that both of these special kernels have efficient implementations: the OSD kernel with a one-element queue and
the EWS kernel with a finite (one-step) state vector and the incremental update,
\[
\bm{Z}^\prime_{t,i}\bm{\kappa} =  \alpha \bm{Z}^\prime_{t-1,i}\bm{\kappa}  + (1-\alpha) \bm{z}^\prime_{t,i} .
\]
We use the OSD kernel for training all models. Modifications of batch norm were required to do time-step-conditional normalization (see Appendix \ref{appendix:training_details}). All experiments use a ResNet-18, which has 8 residual blocks and hence 8 time delays. Note, we conducted experiments with larger ResNets and the additional compute did not affect qualitative properties.
We also add a time delay to the output of the model's first convolutional layer. Consequently, with the OSD kernel, 
the cascaded model requires 9 updates for the output to reach asymptote. The cascaded and serial
models with the same weights will necessarily produce identical asymptotic outputs.


To obtain a finer temporal granularity at evaluation, some simulations switch to the EWS kernel with $\alpha=0.9$.  
Temporal dynamics are qualitatively similar for OSD and EWS, but EWS allows us to better distinguish individual examples 
in terms of their fine-grain timing. EWS with $\alpha=0.9$ requires about 70 steps for the output to asymptote. We note that our findings are robust to the choice of $\alpha$, as long as $\alpha$ slows transmission.

\subsection*{Training Cascaded Networks with TD(\texorpdfstring{$\lambda$}{TEXT})}
\label{subsec:td_lambda_training}


To allow for anytime prediction, we include an output head following each of the $T$ residual blocks in both the serial 
and cascaded models (see Figure~\ref{fig:tdl_resnet}b,c, respectively).
The output heads may share weights or have separate weights.
To encourage correct outputs sooner, we use temporal difference (TD) learning \citep{sutton1988learning} over
the output sequence.  Readers may associate TD methods with reinforcement
learning because TD methods have traditionally been used to predict future rewards. However, TD methods are fundamentally designed for supervised learning.
We use TD to predict a future outcome---the correct classification of an image---from a sequence of successively 
more informative states---the information flowing through the ResNet at each internal time step.

TD($\lambda$) specifies a target output $y_t$ at each time $t \in \{ 1, ... T\}$ based on the model's
actual output $\hat{y}_{t+i}$ at future times $t+i$ for $i>0$, and the eventual outcome or true target, $y_\text{true}$:
\begin{equation}
    y_t = (1-\lambda) \left[ \sum_{i=1}^{T-t}\lambda^{i-1} \hat{y}_{t+i} \right] + \lambda^{T-t}y_\text{true} ,
    \label{eq:td}
\end{equation}
where $\lambda \in [0,1]$ is a free parameter that essentially specifies the time horizon for prediction. TD(1) predicts the eventual
outcome at each step; TD(0) predicts the model's output at the next step (and the eventual outcome at the final step). 
Given target $y_t$ and actual output $\hat{y}_t$, we specify a cross-entropy loss,
$\mathcal{L} = \sum_{t=0}^{T} H(y_t, \hat{y}_t)$, where $H(p,q)$ is the cross-entropy. 
Note that $y_t$ must be treated as a constant, not as a differentiable variable, via
a \texttt{stop\_gradient} (for TensorFlow or Jax) or  \texttt{requires\_grad=False} (for PyTorch).
Although Equation~\ref{eq:td} requires knowledge
of all subsequent network states, the beauty of TD methods is that this loss can be computed incrementally (see Appendix \ref{appendix:TD_experiments}).
The edge cases, TD(0) and TD(1), have trivial implementations. Past research has always used TD(1) for specifying intermediate targets, but we 
will show that TD(1) leads to local optima because the model is penalized for being unable to classify correctly at the earliest steps.

\begin{figure*}[t]
    \centering
    \includegraphics[width=0.9\textwidth]{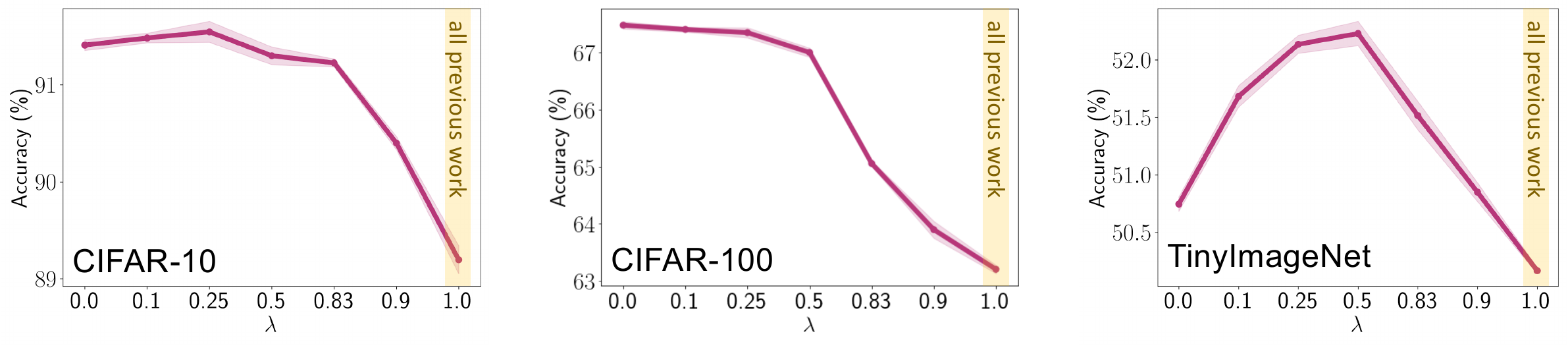}
    \caption{Effect of TD hyperparameter $\lambda$ on \CascadedTD\ test accuracy for three data sets.
    $\lambda=1$ corresponds to the training methodology of all past research on anytime prediction, which
    is inferior to any
    $\lambda < 1$ for all data sets.  Shaded error bands---hard to see on most curves---indicate $\pm 1$ SEM, corrected to remove performance variance due to weight initialization and unrelated to $\lambda$ \cite{masson2003using}. 
    }
    \label{fig:TDlambda}
\end{figure*}

\section*{Results}
\label{sec:experiments}

\subsection*{TD(\texorpdfstring{$\lambda$}{TEXT}) Training}
\label{subsec:training_cascaded_net_params}

We conducted a sweep over hyperparameter $\lambda$ to determine its effect on asymptotic accuracy of \CascadedTD. 
Figure \ref{fig:TDlambda} shows results from five replications of \CascadedTD\ on CIFAR-100, CIFAR-10, and TinyImageNet.
The hyperparameter has a systematic effect, consistent with classic studies with linear models
\citep[][Chapter 12]{sutton2018reinforcement}.
The same effect is observed with high resolution images; see Appendix \ref{appendix:imagenet_results}, where we train a subset of ImageNet. 
Importantly, $\lambda=1$, which is the implicit choice of \emph{every} previous anytime-prediction model
\cite{bolukbasi2017adaptive,Elbayad2020,hu2018anytime,hu2019learning,huang2018,huang2017multi,iuzzolino2019convolutional,kaya2019shallow,Larsson2017,lee2018anytime,marquez2018deep,mcintosh2018recurrent,newell2016stacked,scardapane2020should,teerapittayanon2016branchynet,wang2017idk,yang2020resolution,zamir2017feedback},
obtains the poorest performance for all data sets, significantly worse than $\lambda \approx .5$.
The essential explanation is that larger $\lambda$ penalize the network for behavior it does not have the capability to
achieve: obtaining the asymptotic prediction at the earliest time steps.  
To paraphrase the classic illustration of TD from Sutton \cite{sutton1988learning}, if the 
task is predicting the weather on December 31, no model can predict as accurately on December 1 as on 
December 30. Selecting $\lambda<1$ shortens the prediction horizon; $\lambda=0$ corresponds with requiring a prediction 
only of the weather on the next day. 

In the rest of the article, we report results for \CascadedTD\ with $\lambda=0$, or \emph{TD(0)}. Although TD(0) is not
optimal for all data sets, it is strictly superior to TD(1) and has a trivial implementation because it does not require eligibility traces. Further, it 
avoids the need for a separate validation set to pick $\lambda$.


\subsection*{Anytime Prediction and Speed-Accuracy Trade Offs}
\begin{figure*}[t]
    \centering
    \includegraphics[width=\textwidth]{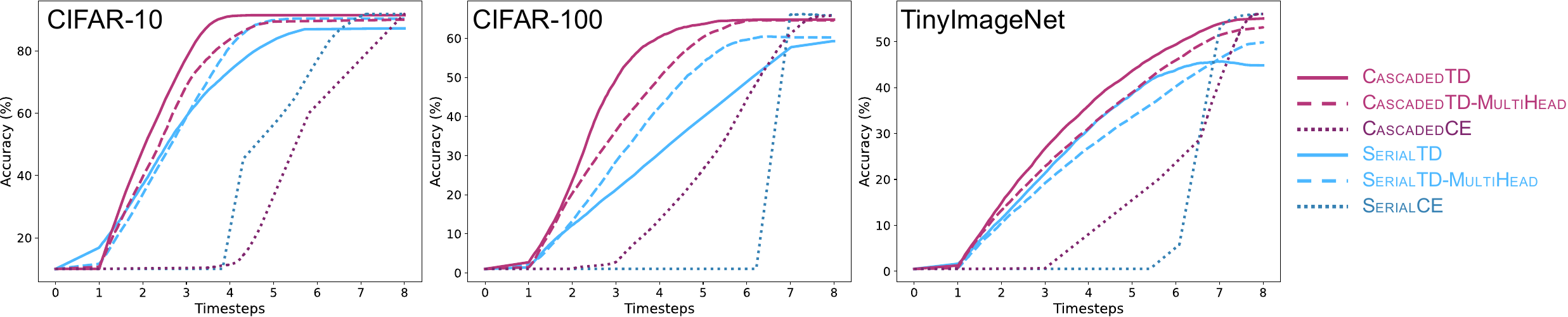}
    \caption{Speed accuracy trade off for three data sets and six models, obtained by varying a stopping threshold and measuring mean latency and mean accuracy. \CascadedTD\ is our parallel anytime prediction model; \SerialTDMH\ is the state-of-the-art method, SDN \cite{kaya2019shallow}.}
    \label{fig:speed-accuracy}
\end{figure*}

Given a static input, an anytime prediction model attempts to obtain the best classification possible as quickly as 
possible. Anytime prediction can be performed by both serial and cascaded models. Both yield
predictions at each time slice, as depicted by the yellow trapezoids in Figures~\ref{fig:tdl_resnet}b,c, which
denote model readout. Critical to anytime prediction is deciding when to terminate processing and initiate a response
\cite{kaya2019shallow,Chen2020,teerapittayanon2016branchynet}.
Following \cite{kaya2019shallow} and \cite{teerapittayanon2016branchynet}, we assume that processing
terminates when the confidence (probability) for the most likely class rises above threshold $\theta$.
For any $\theta$, one can measure the mean stopping time and the mean accuracy for all instances in a test set. By sweeping
$\theta  \in [0,1]$ and plotting mean accuracy as a function of mean stopping time, one obtains a \emph{speed-accuracy 
trade off} curve. Figure~\ref{fig:speed-accuracy} shows curves for models we'll describe next. The horizontal axis indicates number of simulation time steps, which is linearly related to the matched number of operations (multiplies, additions, etc.) performed on simulated parallel hardware for all models. (See Appendix \ref{appendix:compcomplexity}
for further details.)

To evaluate the cascaded model, we compare to a recent state-of-the-art method, the \emph{Shallow-Deep Network (SDN)} \cite{kaya2019shallow}. The SDN has the serial architecture depicted in Figure~\ref{fig:tdl_resnet}b. One critical design
decision was whether to have separate read-out heads at each step (\MultiHead) or a shared read-out head
(\SingleHead), i.e., whether weights are separate or shared. From the perspective of the cascaded model, which
considers the vertical columns of Figure~\ref{fig:tdl_resnet}c to be copies of a network unrolled in time, the
\SingleHead\ approach is natural. The SDN, as a serial model, chose the \MultiHead\ approach. We tested all four 
logical combinations of \{\SerialTD, \CascadedTD\} $\times$ \{\SingleHead, \MultiHead\}. We use
\SerialTD\ and \CascadedTD\ as shorthand for the \SingleHead\ variants, and append \MultiHead\ to the model name for
that version. Additionally, we consider \SerialCE\ and \CascadedCE, which are \SingleHead\ variants trained with 
the standard cross entropy loss that penalizes only asymptotic accuracy and does not explicitly attempt to obtain
a speeded response.

The key observations from Figure~\ref{fig:speed-accuracy}, which shows speed-accuracy trade offs for the six models on three data sets, are as follows.
First, our canonical cascaded model, \CascadedTD, obtains better 
anytime prediction than \SerialTDMH\ (i.e., the architecture of SDN). \CascadedTD\ also achieves higher asymptotic
accuracy; its accuracy matches that of \CascadedCE, a ResNet trained in the standard manner. Thus, cascaded models can 
exploit parallelism to obtain computational benefits in speeded perception without costs in 
accuracy.\footnote{We used $\lambda=1$ to train \SerialTD\ and \SerialTDMH, as was done for \emph{all} previously
proposed serial models. Could SDN and other serial models be improved with $\lambda<1$? In Appendix \ref{appendix:addlresults}, we 
show that the serial model with $\lambda=0$ still does not perform as well as \CascadedTD.}
Second, while \MultiHead\ is superior to \SingleHead\ for serial models, the reverse is true for cascaded models.
This finding is consistent with the cascaded architecture's perspective on anytime prediction as unrolled 
iterative estimation, rather than, as cast in SDN, as distinct read out heads from different layers of the network. 
Third, models trained with TD outperform
models trained with standard cross-entropy loss. Training for speeded responses reorganizes knowledge in the
network so that earlier layers are more effective in classifying instances. We now turn to better understand
what this reorganization entails.

%

\subsection*{Organization of Knowledge in TD-Trained Cascaded Model}

Having examined the response profile of our models over an evaluation set, we now turn to analyzing the response 
to individual instances. Specifically, we ask about the time course of reaching a classification decision.
We define the \emph{selection latency} for an instance to be the minimum number of steps required to 
reach a confidence threshold on one class and maintain that level going forward, i.e.,
$\min \{ t ~|~  [\exists j ~|~ \hat{y}_{t',j} > \theta ~ \forall~ t' \geq t ] \}$, where  $\hat{y}$ is the model output, $j$ is an index over 
classes, and $\theta$ is the threshold. The selection latency does not specify whether or not the chosen class is correct.
We picked a threshold $\theta=0.83$ for \CascadedTD\ such that only $\sim10$\% of the test examples failed to reach threshold; results that follow are robust to this selection.

What determines an instance's selection latency?
Figure~\ref{fig:rapidslow} presents instances that have the lowest and highest latency---labeled `rapid' 
and `slow', respectively. For \CascadedTD\ (Figure~\ref{fig:rapidslow}a), notice the homogeneity of the rapid images:
the objects are viewed from a canonical perspective and lie against a solid background with no clutter in the 
image. In contrast, the slow images are more varied, both in the object's instantiation in the image and the 
background complexity.  Turning to \CascadedCE\ (Figure~\ref{fig:rapidslow}b), instances do not appear to 
stratify by prototypicality. In the rest of this  section, we formalize the notion of prototypicality 
with three measures and compute the correlation of each measure with selection latency for the cascaded model
trained with a TD loss (\CascadedTD) and with the standard cross-entropy loss (\CascadedCE).  Our three measures 
are as follows.


\vspace{-.10in}
\begin{itemize}[leftmargin=*,itemsep=0pt,parsep=2pt]
\item 
\textit{Centrality.}  We compute the cosine distance of an instance's embedding (the penultimate layer activation) and the 
target-class weight vector. The larger this quantity, the better aligned the two vectors are. Because the weight vector will tend to point near
the center of class instances, the cosine distance is a measure of instance centrality.
\item
\textit{C-score.}  Jiang et al.~\cite{jiang2020characterizing} describe an instance-based measure of statistical regularity called the C-score. The C-score 
is an empirical estimate of the probability that a network will generalize correctly to an instance if it is held out from the training set. It
reflects statistical regularity in that an instance similar to many other instances in the training set should have a high C-score.
\item
\textit{Human labeling consistency.}  Peterson et al.~\cite{peterson2019human} collected human labels on images. Most images are consistently labeled, but 
some are ambiguous. The negative entropy of the response distribution indicates inter-human labeling agreement. Presumably, consistently labeled instances are more prototypical.
\end{itemize}
All three measures are available only
for the CIFAR-10 training set. Consequently, we ran 10-fold cross validation on the training set, assessing the correlation
based on the held out images in each fold.  To obtain a granular selection latency, we use the EWS kernel.

Table~\ref{table:kendalls_tau_quant} presents
the correlation---Spearman's $\rho$---between the three prototypicality measures and negative selection latency for
\CascadedCE\ and \CascadedTD.
A positive coefficient indicates shorter latency for prototypical instances.
The coefficient is reliably positive ($p<.001$) for each of the three typicality measures and both models.
However, \CascadedTD\ obtains reliably higher correlations than \CascadedCE\ on two of the three measures ($p<.001$); they are not significantly different on the human consistency measure ($p=.29$). Thus, 
by these quantitative scores, the TD training procedure leads to better stratification of instances 
by typicality, in line with the qualitative results presented in Figure~\ref{fig:rapidslow}.
Why does TD training distinguish instances based on prototypicality? Intuitively, a prototypical instance shares features with many other class instances. Because these features are frequent in the data set, the TD loss focuses on rapidly classifying instances with those features.

\begin{table}[t!]
    \caption{Spearman rank correlations between three measures of instance prototypicality and selection latency for CIFAR-10. Prototypicality measures are formulated such that lower values correspond to higher prototypicality. Since the Spearman coefficient measures the degree to which prototypicality varies with selection latency, large coefficients indicate a correlation between fast responses and the prototypicality of an instances, whereas coefficients close to zero indicate no correlation. Here, we show that the prototypicality and selection latency of an instance for cascaded models are correlated, with \CascadedTD~ yielding significantly higher correlations for two out of three measures as compared to \CascadedCE.}
    \label{table:kendalls_tau_quant}
    \centering
    \small
    \begin{tabular}{|c|c|c|}
        \hline
            & \multicolumn{2}{c|}{\textbf{Spearman's $\rho$}}  \\ \cline{2-3} 
        \multirow{-2}{*}{\textbf{Measure}} & \colorCascadedCE{\CascadedCE} & \colorCascadedTD{\CascadedTD} \\ \hline
        centrality                        & \colorCascadedCE{0.140}       & \highlight{\colorCascadedTD{0.352}} \\ \hline
        C-score                           & \colorCascadedCE{0.326}      & \highlight{\colorCascadedTD{0.489}} \\ \hline
        human consistency                & \highlight{\colorCascadedCE{0.153}}       & \highlight{\colorCascadedTD{0.142}}  \\ \hline
    \end{tabular}
\end{table}

\begin{SCfigure}[10][b!]
    \centering
    \includegraphics[width=.45\textwidth]{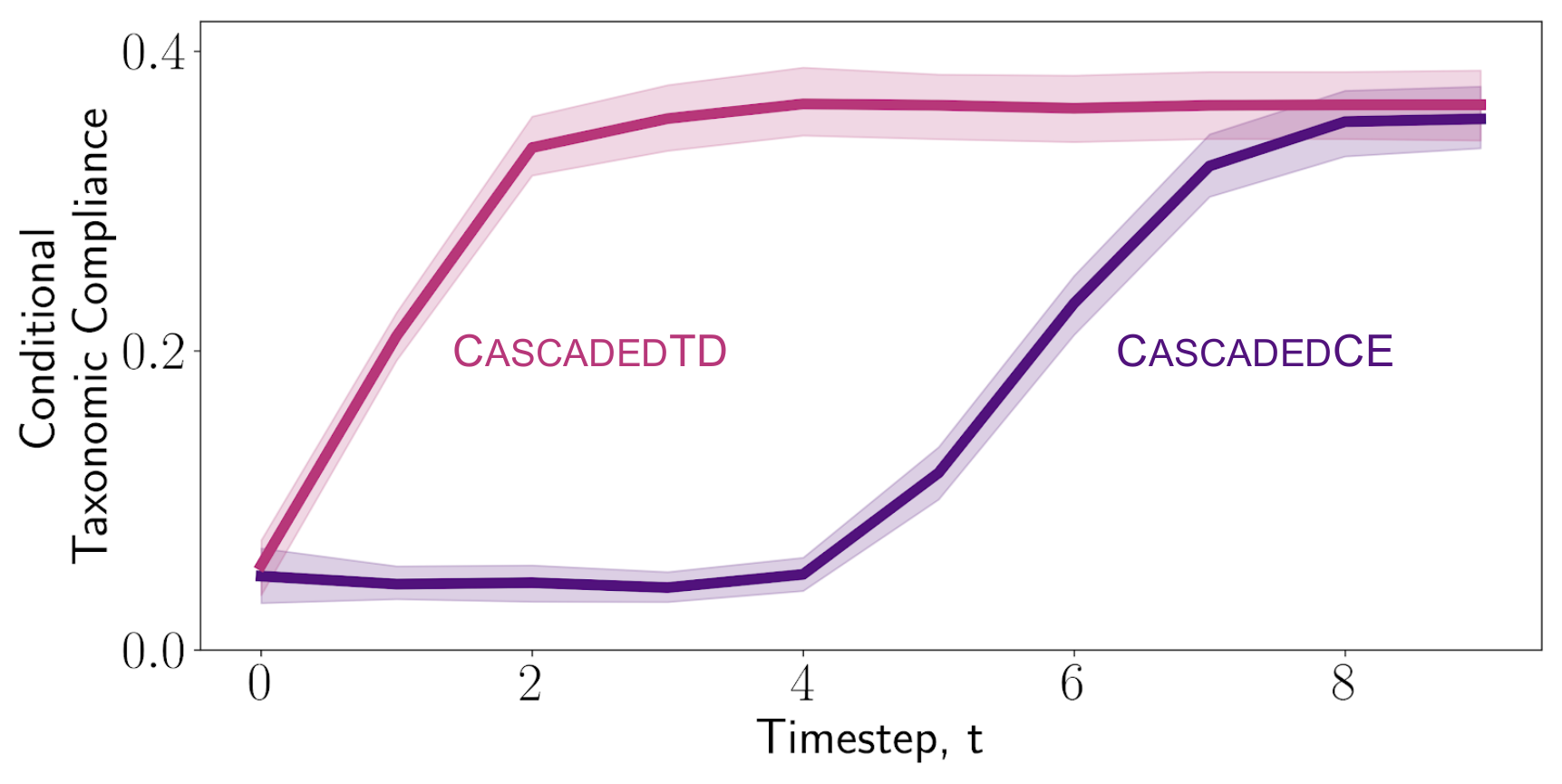}
    \caption{\CascadedTD\ performs coarse-grain classification on CIFAR-100 before fine-train classification, as assessed by a measure of
    conditional taxonomic compliance \cite{zamir2017feedback}. 
     Graphs are based on 5 runs of each model with different random initializations. Shaded 
     error bands indicate $\pm1$ SEM, corrected to remove performance variance due to initial seed \cite{masson2003using}.\newline}
    \label{fig:taxonomic-compliance}
\end{SCfigure}
Beyond investigating the time course of fine-grain classification, we also examined coarse-grain classification.  Forming twenty
superclasses from the 100 fine-grain classes of CIFAR-100, as specified in \cite{krizhevsky2009learning}, we examined the probability of correct
coarse-grain classification conditional on incorrect fine-grain classification. Zamir et al.~\cite{zamir2017feedback} refer to this probability as \emph{taxonomic
compliance}, which reflects information being transmitted about coarse category even when the specific class cannot be determined.
As shown in Figure~\ref{fig:taxonomic-compliance}, taxonomic compliance rises faster for \CascadedTD\ than for \CascadedCE. Whereas chance compliance
is .05, \CascadedTD\ achieves a compliance probability of .35 after 2 steps. \CascadedCE\ requires 8 steps to achieve the same performance.
TD training pushes instances to the correct semantic neighborhood sooner, even when not to the correct class label. This result further supports the reorganization of knowledge for more robust decision making.

\vspace{-.05in}
\subsection*{Robustness to Input Noise} 
\begin{SCfigure}[10][t]
    \centering
    \includegraphics[width=0.54\textwidth]{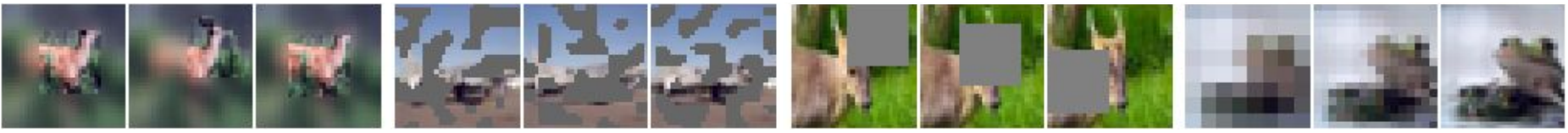}
    \small
    \caption{Lossy noise types. From left to right: \textit{Focus}, \textit{Perlin}, 
    \textit{Occlusion}, \textit{Resolution}.}
    \label{fig:stationary_noise_types4}
\end{SCfigure}

\vspace{-.04in}
\label{subsec:noise_robustness}
In previous simulations, we have assumed the input was static while internal processing took place.
Now we consider static inputs with time-varying noise. 
Figure~\ref{fig:stationary_noise_types4} shows four types of lossy noise we consider on CIFAR-10 
images.  The noise types are:
(1) \textit{Focus}: a $16 \times 16$ foveated patch randomly placed within the image, where regions outside of the patch are Gaussian blurred;
(2) \textit{Perlin}: gradient noise randomly applied to 40\% of image pixels;
(3) \textit{Occlusion}: a $16 \times 16$ occluding patch randomly placed within the image; 
(4) \textit{Resolution}: random downsampling by factors of
$2\times$ or $4\times$ 
via average pooling followed by $k$-nearest upsampling to recover the original dimensionality of $32 \times 32$. 

For each noise type, \CascadedCE\ and \CascadedTD\ models are trained with the corresponding image transformation type as a data augmentation.
Because the external environment changes more rapidly than any snapshot of the environment can be 
processed, cascaded models will necessarily integrate signals from multiple snapshots. To determine
whether signal integration is beneficial for noise suppression, we compared to a serial model that is
allowed to fully process each snapshot, which for the architecture requires $9\times$ as many 
sequential updates as the cascaded model. We therefore refer to the model as \Sequential; it 
has the same weights as \CascadedCE.

Two noise variants are applied to test images: \emph{persistent}, in which
a noise sample is drawn independently at each update step, and \emph{transient}, which consists of three stages: first, the model reaches
its asymptotic output on a noise-free input; second, the input is corrupted by noise samples for a variable number
of steps; and third, the noise-free input is presented until the model returns to its previous 
asymptotic output. For persistent noise, we assess with asymptotic accuracy; for transient noise, we 
assess with a measure of \emph{drop in integrated performance} over the course of the episode,
which indicates how quickly the model recovers from noise perturbation (smaller is better).
Simulation details provided in Appendix \ref{appendix:noise_experiment_details}.

\setlength\tabcolsep{2.5pt}
\begin{table}[b]
    \small
    \centering
    \caption{Experiments on persistent and transient input noise applied to CIFAR-10. Highlight indicates the best performance.}

    \label{table:noise_results}
    \begin{tabular}{c|c|c|c||c|c|c|}
        \cline{2-7}
        & \multicolumn{3}{c||}{\textbf{Persistent Noise}}  & \multicolumn{3}{c|}{\textbf{Transient Noise}} \\
        & \multicolumn{3}{c||}{\textit{Asymptotic Accuracy (\%)}}  & \multicolumn{3}{c|}{\textit{Drop in Integrated Performance}}  \\ \hline
        \multicolumn{1}{|c|}{\textbf{Noise}} & \colorSequential{\scriptsize \sc SerialCE$\times$9} & \colorCascadedCE{\scriptsize \sc CascadedCE} & \colorCascadedTD{\scriptsize \sc CascadedTD} 
        & \colorSequential{\scriptsize \sc SerialCE$\times$9} & \colorCascadedCE{\scriptsize \sc CascadedCE} & \colorCascadedTD{\scriptsize \sc CascadedTD} \\ \hline
        \multicolumn{1}{|c|}{Focus} & \colorSequential{84.27 $\pm$ 0.06} & \colorCascadedCE{83.75 $\pm$ 0.10} & \highlight{\colorCascadedTD{87.31 $\pm$ 0.04}} & \colorSequential{0.62 $\pm$ 0.04} & \colorCascadedCE{0.66 $\pm$ 0.05} & \highlight{\colorCascadedTD{0.00 $\pm$ 0.01}}\\ \hline
        \multicolumn{1}{|c|}{Occlusion} & \colorSequential{86.26 $\pm$ 0.08} & \colorCascadedCE{82.73 $\pm$ 0.09} & \highlight{\colorCascadedTD{89.76 $\pm$ 0.05}}  & \colorSequential{7.70 $\pm$ 0.55} & \colorCascadedCE{8.25 $\pm$ 0.72} & \highlight{\colorCascadedTD{0.93 $\pm$ 0.15}} \\ \hline
        \multicolumn{1}{|c|}{Perlin} & \colorSequential{85.18 $\pm$ 0.03} & \colorCascadedCE{84.56 $\pm$ 0.05} & \highlight{\colorCascadedTD{87.67 $\pm$ 0.08}}  & \colorSequential{0.86 $\pm$ 0.06} & \colorCascadedCE{0.87 $\pm$ 0.07} & \highlight{\colorCascadedTD{0.00 $\pm$ 0.01}}\\ \hline
        \multicolumn{1}{|c|}{Resolution} & \colorSequential{84.53 $\pm$ 0.07} & \colorCascadedCE{85.40 $\pm$ 0.07} & \highlight{\colorCascadedTD{88.19 $\pm$ 0.10}}  & \colorSequential{0.81 $\pm$ 0.06} & \colorCascadedCE{0.53 $\pm$ 0.05} & \highlight{\colorCascadedTD{0.18 $\pm$ 0.02}} \\ \hline
    \end{tabular}
\end{table}

Table~\ref{table:noise_results} indicates that \CascadedTD\ obtains a degree of robustness to
persistent and transient noise not matched by the alternative models. Although \Sequential\ performs a 
full inference pass  on each noise perturbation, the stateful nature of \CascadedTD\ allows it to smooth 
out noise via slow integration. Although \CascadedCE\ shares the same architecture as \CascadedTD,
TD training is required to orchestrate the integration of content-specific perceptual information.
This experiment indicates that laggy information processing in the cascaded model can be advantageous
in a noisy environment. In Appendix \ref{appendix:noise_experiment_details}, we note that the benefit for \CascadedTD\ applies only to lossy noise,
not translations and rotations, as one would expect from the perspective of noise averaging.

\vspace{-.02in}
\subsection*{Meta-cognitive Inference}
\vspace{-.01in}
\label{subsec:metacog}
In this section, we consider the hypothesis that temporally intermediate outputs from cascaded networks can provide additional signals 
to improve performance. We term this \emph{metacognition}, by reference to human abilities to reason about our reasoning processes. 

\label{subsec:ood_metacog}
The temporal trace of output from \CascadedTD\ is provided to a separate classifier, \MetaCog,
which is \emph{discriminatively} trained for out-of-distribution (OOD) detection.
\MetaCog\ is a fully connected feedforward net with a 256-unit hidden layer and a 
sigmoidal output unit for binary prediction: 1 or 0 for in- or out-of-distribution instances, respectively. CIFAR-10's validation set serves as the in-distribution training examples, whereas the validation sets of TinyImageNet, LSUN, and SVHN serve as OOD training examples; see details in Appendix \ref{appendix:ood_dataset_details}.
The \CascadedTD\ output is represented in one of four ways as input to \MetaCog: 
(1) the confidence of the most probable class, known as the \emph{max softmax prediction} (\emph{MSP}), 
(2) entropy of the class posterior distribution, 
(3) the class posterior distribution, and
(4) the logit representation of the posterior.
We investigate whether feeding the output of all time steps to \MetaCog\ leads to improved prediction relative to feeding only the final asymptotic output. Only the latter information is available in a standard feedforward 
net.

\begin{SCfigure}[10][tb]
    \centering
    \includegraphics[width=0.445\textwidth]{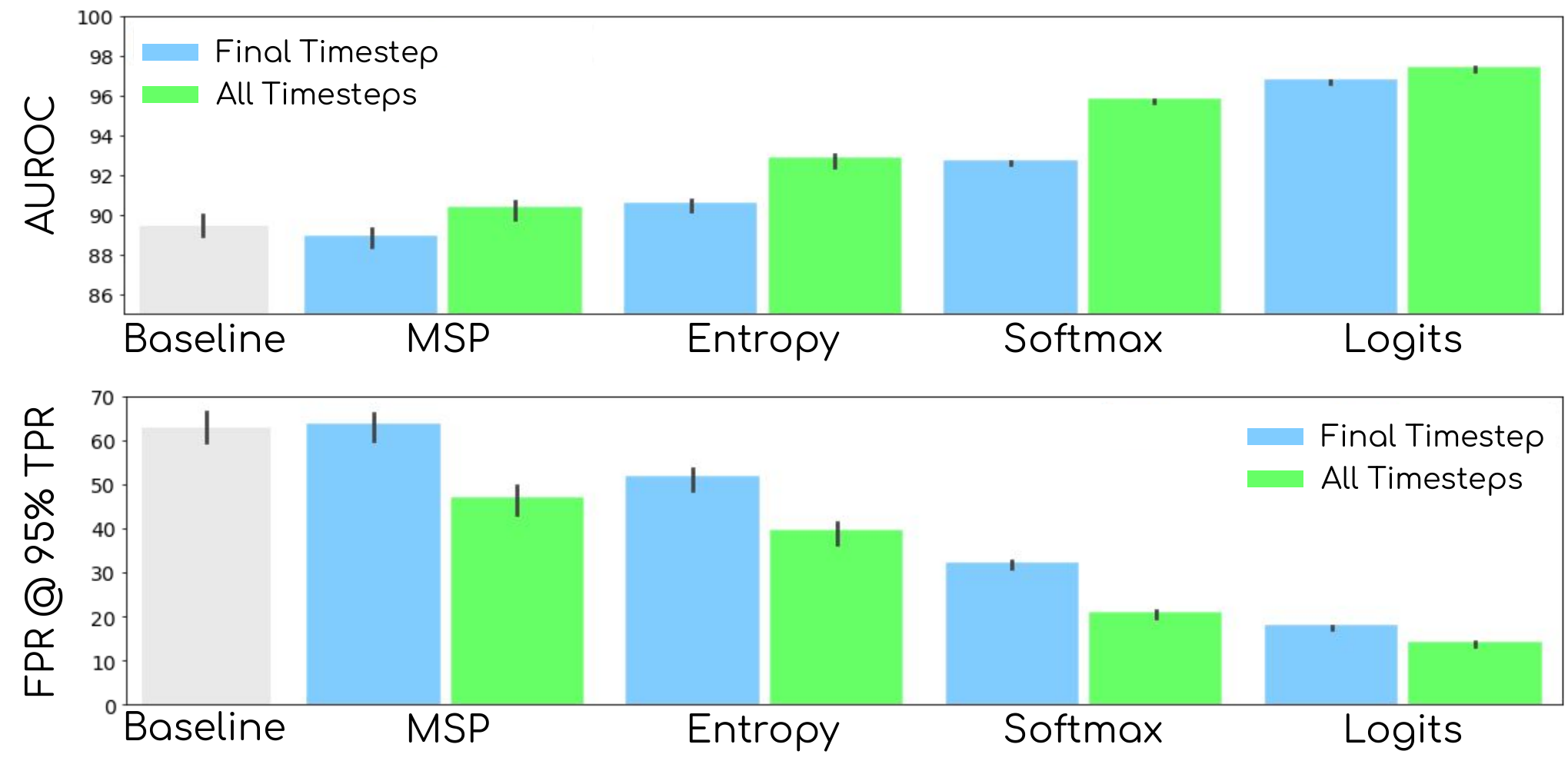}
    \small
    \caption{On CIFAR-10, the output of \CascadedTD\  over time provides a reliable signal for improving OOD
    detection over using just its asymptotic output. Larger AUROC and smaller FPR are better. All output
    representations benefit from the temporal trajectory. 
    Error bars reflect $\pm 1$ SEM corrected to remove comparison-unrelated variance \cite{masson2003using}. 
    Baseline is the final max-softmax prediction.}
    \label{fig:metacog_ood}
\end{SCfigure}

Following \cite{liang2017enhancing}, we assess OOD performance with \textit{AUROC}, the area under the 
ROC curve, and \textit{FPR @ 95\% TPR}, the false positive rate at 
95\% true positive rate. A baseline metric is computed directly from the final max softmax predictions of the \CascadedTD\ model, whereas the other metrics are based on the \MetaCog\ model output.
Figure~\ref{fig:metacog_ood} indicates that the temporal output trajectory of \CascadedTD\ provides a valuable
signal for OOD detection. Our goal here was not to propose a method for OOD detection, but merely to 
demonstrate that in principle, there is information about the input that is conveyed by the cascaded model's
dynamics but that is not available in a traditional classifier's output.



\vspace{-.02in}
\section*{Discussion}
\vspace{-.01in}
\label{sec:conclusion}
We investigated a neglected biologically-motivated architecture in which the bottleneck in neural information processing is transmission delays, 
not the number of neurons that can update in parallel.
We proposed a temporal-difference (TD) loss that yields improved speed-accuracy trade offs.
We showed that this model beats the state-of-the-art anytime prediction method, partly because
of the TD loss and partly because of the model dynamics.  The cascaded model has many distinctive 
properties, including: it classifies prototypical instances more rapidly than outliers; it performs 
coarse-to-fine semantic processing in which general semantic categories are rapidly inferred even when 
specific class labels are not; it is able to moderate time-varying input noise; and the temporal trace 
of the model's output provides an additional signal that can be exploited to improve information 
processing beyond that provided by the asymptotic model output.
Of course, these interesting properties come at a computational cost when parallel hardware is simulated
on existing compute infrastructure.
We see three directions in which cascaded nets have particular potential.
\begin{itemize}[leftmargin=*,itemsep=0pt,parsep=3pt]
\vspace{-2.5mm}
\item
For neuroscientists using deep nets as a model of human vision, cascaded nets are a better approximation to the dynamics of the neural hardware.
The properties we investigate---neurons operate in parallel, neurons are stateful, and neurons are slow to transmit information---seem likely to have 
a critical impact on the  nature of cortical computing. As one simple illustration, cortical feedback processes are often posited to be critical 
for explaining differences in processing efficiency of visual stimuli \citep[e.g.,][]{Kar2019,Spoerer2017}.
We have shown that these difference might be partly explained 
by feedforward cascaded dynamics.
\item
For hardware researchers, cascaded networks are a possible direction for the future design of AI hardware. It is a direction quite unlike modern GPUs and TPUs,
one that exploits massively parallel albeit slow and possibly noisy information processing. Our success in showing strong performance from cascaded models,
as well as a training procedure to obtain quick and accurate responses, should encourage research in this direction.
\item
For AI research in anytime prediction, we've shown that existing models can be improved with a TD($\lambda$) loss; all past research has adopted $\lambda=1$, which we show to be inferior to $\lambda<1$. For researchers who
care little about cascaded models per se,  cascaded models offer an intriguing method to train serial
feedforward models. One can take a serial feedforward model, turn it into a cascaded model for training 
with TD methods, and then run it in serial mode. We've shown that TD training can improve asymptotic 
model accuracy while still providing anytime predictions due to inductive biases it imposes on the 
organization of representations. 

\end{itemize}

\newpage


\nocite{deng2009imagenet}
\nocite{liang2017enhancing}
\nocite{masson2003using}
\nocite{netzer2011reading}
\nocite{yu2015lsun}

\section*{Acknowledgments}

The authors thank Tyler Scott, Anirudh Goyal, 
Pradeep Shenoy, and particularly our three anonymous reviewers for thoughtful comments and feedback on earlier drafts of this work.

\bibliography{references}
\bibliographystyle{plainnat}

\appendix
\counterwithin{figure}{section}
\setcounter{table}{0}
\renewcommand{\thetable}{\Alph{section}.\arabic{table}}

\section{Experiment Details}
\label{appendix:training_details}

\subsection{\CascadedCE~and \CascadedTD~Experiment Details}
For all cascaded and serial models, we used a ResNet-18 for CIFAR-10, CIFAR-100, and TinyImageNet datasets.  We experimented with deeper nets, up to ResNet-52, but found no differences in model behavior.
The models were trained using data parallelism over 8 GPUs (see \S \ref{appendix:comp_infra} for infrastructure details), with the model on each GPU using a batch size of 128. SGD with Nesterov momentum, an initial learning rate of 0.1, weight decay of 0.005, and momentum of 0.9 was used to optimize a softmax cross-entropy loss for \SerialCE/\CascadedCE\ and a temporal difference cross-entropy loss for \SerialTD/\CascadedTD. All models were trained for 120 epochs and the learning rate was decayed with a multiplicative factor of 0.2 every 30 epochs. 

The training datasets were split (class-balanced) as 90-10 train-validation, where the validation splits were held out for downstream tasks, such as training \MetaCog\ models (see Appendix \ref{appendix:subsec:metacog}). For \CascadedTD, the batch normalization layer must be augmented such that running means and variances are tracked independently for each timestep. At run-time, if the maximum number of timesteps used during training is exceeded, as occurs when using the EWS kernel, the final timestep statistics of the batch normalization layers are used for all subsequent timesteps. Furthermore, we observed that the offset parameter of the affine transformation of the batch normalization on identity mappings blows up to NaN values during training; consequently, we do not use batch normalization on the identity mapping in the cascaded nor serial models.

\subsection{Temporal Difference Loss}
\subsubsection{Incremental TD Formulation}
\label{appendix:TD_experiments}
TD($\lambda$) amounts to training at each time step $t$ with a target, $y_t^\lambda$, that is an exponentially decaying trace of future outputs, anchored beyond 
some asymptotic time $T$ to the true target, $y$. Denoting the network output at step $t \in \{1, \ldots, T \}$  as $\hat{y}_t$, the trace is:
\begin{equation*}
    \begin{split}
    y_t^\lambda &= (1 - \lambda) \sum_{n=1}^{\infty}  \lambda^{n-1} \bar{y}_{t+n} \\
    &\textrm{~~~~~~~~with~ } \bar{y}_{t+n} = \begin{cases} \hat{y}_{t+n} \text{~~~if } t + n \leq T\\ y \text{~~~~~~~~~otherwise} \end{cases}\\
    &= (1 - \lambda) \sum_{n=1}^{T-t} \lambda^{n-1} \hat{y}_{t+n} + \lambda^{T-t} y .
    \end{split}
\end{equation*}

We train with cross entropy loss over all time steps. For a single example, the loss is
\begin{equation*}
    \mathcal{L} = - \sum_{t,i} y^\lambda_{ti} \ln \bar{y}_{ti}.
\end{equation*}
The derivative of this loss with respect to the network parameters $w$ can be expressed in terms of the derivative with respect to the logits:
\begin{equation*}
    \nabla_w \mathcal{L} = -\sum_{t,i} (y^{\lambda}_{ti} - \bar{y}_{ti}) \nabla_w z_{ti} ,
\end{equation*}
where $z_{ti}$ is the logit of class $i$ at step $t$.
The temporal difference method provides a means of computing this gradient incrementally, such that at each step $t$, an update can be computed based on only the 
difference of model outputs at $t$ and $t+1$:
\begin{equation*}
    \nabla_w^\textsc{td} \mathcal{L} = -\sum_{t,i} (\bar{y}_{t+1,i} - \bar{y}_{ti}) e_{ti},
\end{equation*}
where $e_{ti}$ is an \emph{eligibility trace}, defined as:
\begin{equation*}
    e_{ti} = \begin{cases} 
    \boldsymbol{0} & \textrm{if } t = 0 \\
    \lambda e_{t-1,i} + \nabla_w z_{ti} & \textrm{if } t\ge 1 
    \end{cases}
\end{equation*}

The incremental formulation of TD via $\nabla_w^\textsc{td} \mathcal{L}$ is valuable when gradients and/or weight updates must be computed 
on line rather than presenting an entire sequence before computing the loss, e.g., in the situation where the network runs for many steps and 
truncated BPTT is required. In our experiments, we use the summed gradient, $\nabla_w \mathcal{L}$, computed by PyTorch from the full $T$ step sequence
and our exponentially weighted target, $y^\lambda_t$.

\subsubsection{TD(\texorpdfstring{$\lambda$}{TEXT}) Sweep} 

Table \ref{table:td_lambda_experiment_results} shows the tabulated results for asymptotic accuracy of \CascadedTD\ swept over $\lambda$ values on CIFAR-10, CIFAR-100, and TinyImageNet, as well as \CascadedCE. Note, 5 trials per $\lambda$ were trained for each dataset.

\subsection{Data Augmentation}
\label{appendix:cascade_data_augmentation_details}
When training all models on CIFAR-10 and CIFAR-100, for each batch the $32 \times 32$ images are padded with 4 pixels to each border (via reflection padding), resulting in a $40 \times 40$ image. A random $32 \times 32$ crop is taken, the image is randomly flipped horizontally, and standard normalized using the training set statistics is applied. Finally, a random $8 \times 8$ block cut is taken such that the cropped pixels are set to 0. Images at run-time are only standard normalized using training set statistics - no other augmentation is applied with the exception of the persistent and noise robustness experiments. The same process is followed for TinyImageNet with the following exceptions: (1) the $64 \times 64$ images are padded to $86 \times 86$ with reflection padding, random cropped back to $64 \times 64$, randomly flipped horizontally, then standard normalized, and (2) no $8 \times 8$ block cutting is applied.

\captionsetup{font={footnotesize}}
\begin{table*}[!]
    \vskip -0.15in
    \caption{Asymptotic accuracy for \CascadedTD\ models for various $\lambda$, as well as \CascadedCE.
    \textcolor{bestGreen}{Green font} indicates best performance across TD($\lambda$) and \CascadedCE\ models for a given dataset. Highlight indicates best performing TD($\lambda$) across $\lambda$'s for a given dataset.}
    \label{table:td_lambda_experiment_results}
    \vskip 0.15in
    \centering
     \fontsize{7.}{10}\selectfont
    \begin{tabular}{c|c|c|c|c|c|c|}
    \cline{2-7}
    \multicolumn{1}{l|}{} & \multicolumn{6}{c|}{\textbf{Cascaded Model Variant}} \\ \hline
    \multicolumn{1}{|c|}{\textbf{Dataset}} & \sc \textbf{TD(0)} & \sc \textbf{TD(0.25)} & \sc \textbf{TD(0.5)} & \sc \textbf{TD(0.83)} & \sc \textbf{TD(1)} & \sc \textbf{CE} \\ \hline
    \multicolumn{1}{|c|}{CIFAR-10} & 91.22 $\pm$ 0.18 & \highlight{91.65 $\pm$ 0.08} & 91.45 $\pm$ 0.16 & 90.98 $\pm$ 0.21 & 88.75 $\pm$ 0.42 & \textcolor{bestGreen}{91.91 $\pm$ 0.08} \\ \hline
    \multicolumn{1}{|c|}{CIFAR-100} & \textcolor{bestGreen}{\highlight{67.48 $\pm$ 0.14}} & 67.35 $\pm$ 0.20 & 67.00 $\pm$ 0.18 & 65.06 $\pm$ 0.11 & 63.20 $\pm$ 0.14 & 65.56 $\pm$ 0.06 \\ \hline
    \multicolumn{1}{|c|}{TinyImageNet} & 50.74 $\pm$ 0.11 & 52.03 $\pm$ 0.07 & \highlight{52.25 $\pm$ 0.07} & 51.39 $\pm$ 0.13 & 49.86 $\pm$ 0.15 & \textcolor{bestGreen}{52.46 $\pm$ 0.06} \\ \hline
    \end{tabular}
    \vskip -0.15in
\end{table*}

\subsection{Noise Experiments}
\label{appendix:noise_experiment_details}
The four noise perturbations studied in the main article are lossy.
Here we consider two additional noise sources that are roughly information preserving:
\textit{Translation}: random shifts $\pm 8$ pixels in $(x,y)$ on a reflection-padded image; 
\textit{Rotation}: random rotations $\pm 60^\circ$ on a reflection padded image. 

\Sequential\ is best on information preserving 
transformations such as \textit{Translation} and \textit{Rotation} because it is performing a full inference pass on the input whereas the cascaded
models are performing a single update step.

While both \CascadedTD\ and \CascadedCE\ smooth responses over frames, \CascadedTD\ performs 
better, indicating that beyond smoothing, TD training orchestrates the integration of image-specific perceptual information.
This integration matters more for lossy transformations, where information integration is essential.

The training details are the same as previous simulations, 
except that  we discard the $8 \times 8$ block data augmentation in  order to avoid biasing the models 
toward the \textit{Occlusion} noise transformation.  We evaluate the cascaded models with the OSD kernel 
to allow for a comparison of cascaded models with \Sequential.

In the persistent-noise experiment, five trials are run per image in the test set.
In the transient-noise experiment, we present the noise-free input for 10 time steps 
(sufficient for the cascaded models to reach asymptote), apply one of the six noise transforms for 
$N$  time steps, and then present the noise-free input for another 10 steps, allowing the model to return 
to its  asymptotic state. We run five trials per condition for each $N \in \{1, 2, 3, 4, 5, 6\}$ and
each image in the test set. Performance is evaluated as the drop-in-integrated-performance, 
$\text{DIP} = \hat{y}_T - \mathbb{E}_{t \in \{B, \ldots, T\}} [\hat{y}_t]$,
where $T$ is the total time steps in the simulation, $B$ is the onset time of the noise transformations, 
and $\hat{y}_t$ is  the model's target-class confidence at time step $t$. DIP indicates how quickly a model can recover from noise perturbations. 

\begin{SCfigure}[10][t]
    \centering
    \includegraphics[width=0.45\textwidth]{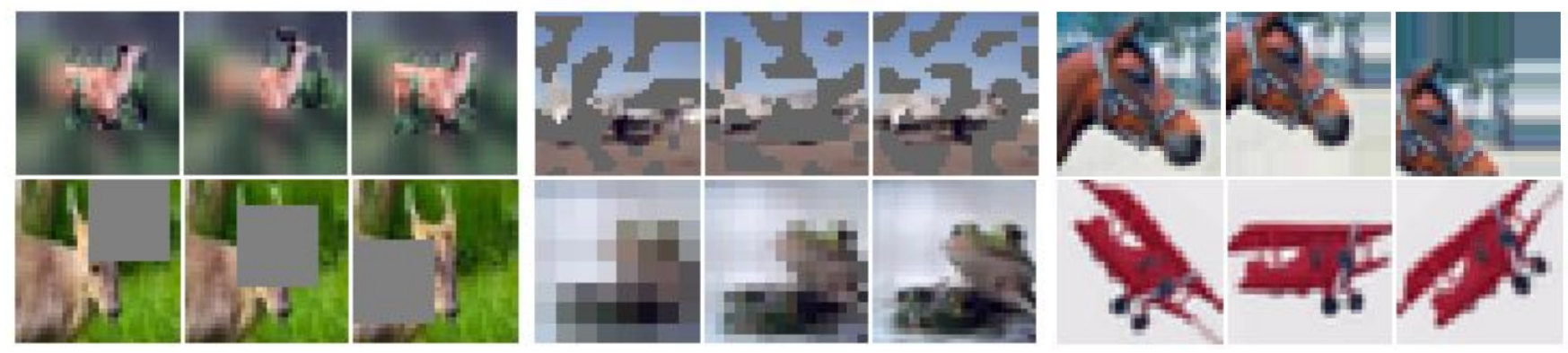}
    \small
    \caption{Noise types. From left to right, top to bottom: \textit{Focus}, \textit{Perlin}, \textit{Translation},
    \textit{Occlusion}, \textit{Resolution}, \textit{Rotation}.}
    \label{fig:stationary_noise_types}
\end{SCfigure}
\setlength\tabcolsep{2.5pt}
\begin{table}[tb!]
    \small
    \centering
    \caption{Persistent-noise experiment. Highlight indicates best asymptotic performance for a given noise type.}

    \label{table:stationary_results_appendix}
    \begin{tabular}{c|c|c|c|}
        \cline{2-4}
        Persistent & \multicolumn{3}{c|}{\textbf{Asymptotic Model Performance}} \\ \hline
        \multicolumn{1}{|c|}{\textbf{Noise}} & \colorSequential{\scriptsize \sc SerialCEx9} & \colorCascadedCE{\scriptsize \sc CascadedCE} & \colorCascadedTD{\scriptsize \sc CascadedTD} \\ \hline
        \multicolumn{1}{|c|}{Focus} & \colorSequential{84.27 $\pm$ 0.06} & \colorCascadedCE{83.75 $\pm$ 0.10} & \highlight{\colorCascadedTD{87.31 $\pm$ 0.04}} \\ \hline
        \multicolumn{1}{|c|}{Occlusion} & \colorSequential{86.26 $\pm$ 0.08} & \colorCascadedCE{82.73 $\pm$ 0.09} & \highlight{\colorCascadedTD{89.76 $\pm$ 0.05}} \\ \hline
        \multicolumn{1}{|c|}{Perlin} & \colorSequential{85.18 $\pm$ 0.03} & \colorCascadedCE{84.56 $\pm$ 0.05} & \highlight{\colorCascadedTD{87.67 $\pm$ 0.08}} \\ \hline
        \multicolumn{1}{|c|}{Resolution} & \colorSequential{84.53 $\pm$ 0.07} & \colorCascadedCE{85.40 $\pm$ 0.07} & \highlight{\colorCascadedTD{88.19 $\pm$ 0.10}} \\ \hline
        \multicolumn{1}{|c|}{Rotation} & \highlight{\colorSequential{89.11 $\pm$ 0.04}} & \colorCascadedCE{73.79 $\pm$ 0.10} & \colorCascadedTD{87.51 $\pm$ 0.03} \\ \hline
        \multicolumn{1}{|c|}{Translation} & \highlight{\colorSequential{87.55 $\pm$ 0.12}} & \colorCascadedCE{76.72 $\pm$ 0.09} & \colorCascadedTD{83.42 $\pm$ 0.14} \\ \hline
    \end{tabular}
\end{table}

\begin{table}[tb!]
    \small
    \centering
    \caption{Transient-noise experiment. Highlight indicates lowest DIP for a given noise type.}
    \label{table:non_stationary_results}
    \begin{tabular}{c|c|c|c|}
        \cline{2-4}
        Transient & \multicolumn{3}{c|}{\textbf{Drop in Integrated Performance (DIP)}} \\ \hline
        \multicolumn{1}{|c|}{\textbf{Noise}} & \colorSequential{\scriptsize \sc SerialCEx9} & \colorCascadedCE{\scriptsize \sc CascadedCE} & \colorCascadedTD{\scriptsize \sc CascadedTD} \\ \hline
        \multicolumn{1}{|c|}{Focus} & \colorSequential{0.62 $\pm$ 0.04} & \colorCascadedCE{0.66 $\pm$ 0.05} & \highlight{\colorCascadedTD{0.00 $\pm$ 0.01}} \\ \hline
        \multicolumn{1}{|c|}{Occlusion} & \colorSequential{7.70 $\pm$ 0.55} & \colorCascadedCE{8.25 $\pm$ 0.72} & \highlight{\colorCascadedTD{0.93 $\pm$ 0.15}} \\ \hline
        \multicolumn{1}{|c|}{Perlin} & \colorSequential{0.86 $\pm$ 0.06} & \colorCascadedCE{0.87 $\pm$ 0.07} & \highlight{\colorCascadedTD{0.00 $\pm$ 0.01}} \\ \hline
        \multicolumn{1}{|c|}{Resolution} & \colorSequential{0.81 $\pm$ 0.06} & \colorCascadedCE{0.53 $\pm$ 0.05} & \highlight{\colorCascadedTD{0.18 $\pm$ 0.02}} \\ \hline
        \multicolumn{1}{|c|}{Rotation} & \colorSequential{0.24 $\pm$ 0.02} & \colorCascadedCE{4.12 $\pm$ 0.29} & \highlight{\colorCascadedTD{0.00 $\pm$ 0.01}} \\ \hline
        \multicolumn{1}{|c|}{Translation} & \highlight{\colorSequential{0.72 $\pm$ 0.05}} & \colorCascadedCE{4.17 $\pm$ 0.37} & \colorCascadedTD{1.53 $\pm$ 0.14} \\ \hline
    \end{tabular}
\end{table}

\subsection{Additional Temporal Dynamics Results}
\subsubsection{Deadline-based stopping criterion}
\begin{figure}[b!]
    \centering
    \includegraphics[width=\textwidth]{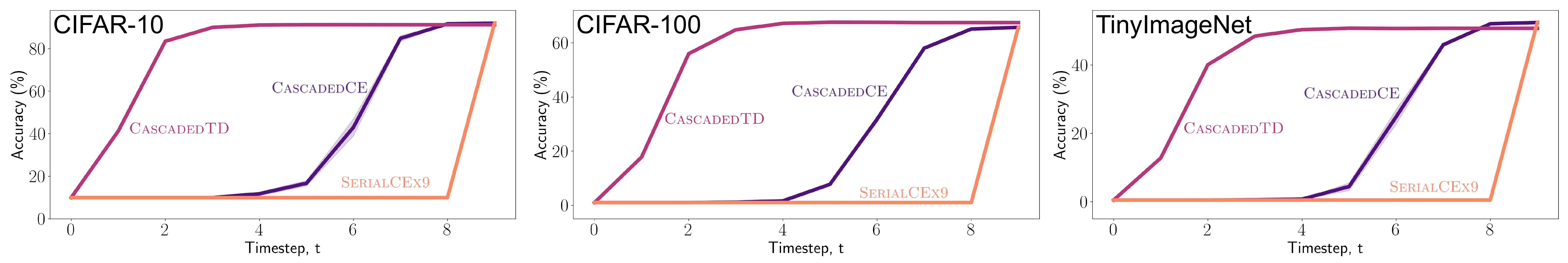}
    \small
    \caption{Speed accuracy trade off for CIFAR-10, CIFAR-100, and TinyImageNet. Here we show the dynamics for a temporal-deadline stopping criterion, whereas Figures \ref{fig:speed-accuracy} and \ref{fig:SerialTD01} show speed accuracy trade offs obtained by a confidence threshold-based criterion. 
    Accuracy assuming a particular stopping time is computed for each daedline. Note, \Sequential\ produces an output only after~only all updates. Confidence intervals across model replications are shown by the shaded regions, which are difficult to see because the uncertainty is small.}
    \label{fig:speed_acc_tradeoff_gt}
    \vskip -0.1in
\end{figure}

In the main paper, we show speed-accuracy trade offs for models based on a stopping criterion that terminates processing when a \emph{confidence threshold} is reached for one output class. In Figure~\ref{fig:speed_acc_tradeoff_gt},
we examine an alternative stopping criterion that is based on a \emph{temporal deadline}, i.e., after a certain number of update iterations. When the speed-accuracy curves for the two stopping criteria are directly compared, the confidence-threshold procedure is superior for all models. For this reason, we report the confidence-threshold procedure in the main paper. However, the confidence-threshold procedure does not allow us to readily compute error bars across model replications because the mean stopping time is slightly different for each replication. In Figure~\ref{fig:speed_acc_tradeoff_gt}, we show confidence intervals at the various stopping times using shaded regions. (The regions are very small and are difficult to see.) The main reason for presenting these curves is to convince readers of the reliability of the speed-accuracy curves.

\subsubsection{Serial models trained with TD}
\label{appendix:addlresults}
\begin{figure}[t]
    \centering
    \includegraphics[width=\textwidth]{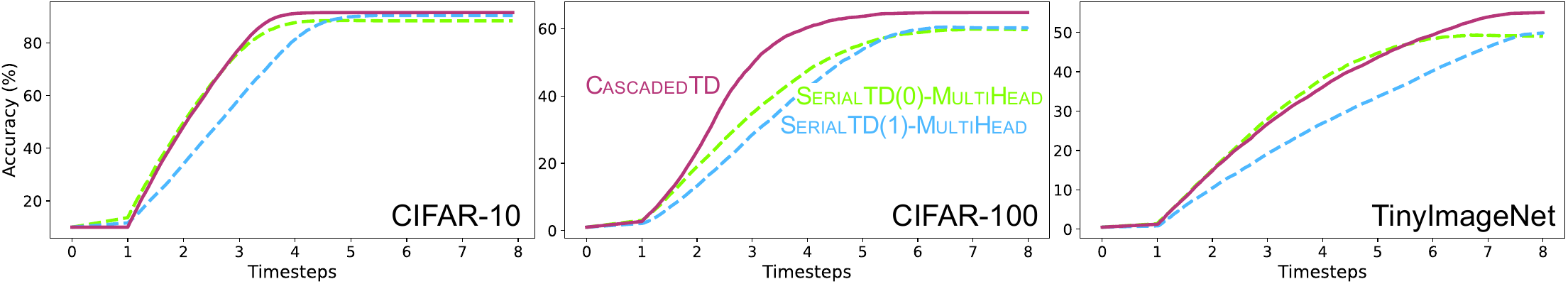}
    \small
    \caption{Comparison of \CascadedTD, 
    {\footnotesize \sc \colorSerialTD{SerialTD(1)-MultiHead}} (Shallow-Deep Nets), and
    {\footnotesize \sc \colorSerialTDZ{SerialTD(0)-MultiHead}}. Shallow-Deep Nets can be improved using TD(0), but
    the speed-accuracy trade off is still significantly worse on CIFAR-100, and asymptotic accuracy is lower than that
    of \CascadedTD.}
    \label{fig:SerialTD01}
\end{figure}
In the main paper, we compare cascaded models to Shallow-Deep Networks, 
a serial model with multi-headed outputs trained with TD(1). Just as training
with $\lambda<1$ improves performance of \CascadedTD, one might hope to
observe a similar benefit for serial models such as Shallow-Deep Networks. 
Figure~\ref{fig:SerialTD01} shows that indeed training with TD(0) is
superior to training with TD(1) for SDNs, labeled in the graph as
{\footnotesize \sc \colorSerialTD{SerialTD-MultiHead}}. 
The serial model's performance improves significantly, nearly to the
level of \CascadedTD, for two data sets, but for the third, 
\CascadedTD\ still has a  considerable advantage over the serial 
model, whether trained with  $\lambda=0$ or $\lambda=1$.

\subsubsection{Qualitative performance of TD trained models on CIFAR-10}
\begin{figure}[h]
    \centering
    \begin{minipage}{.35\textwidth}
      \centering
      \includegraphics[width=.94\linewidth]{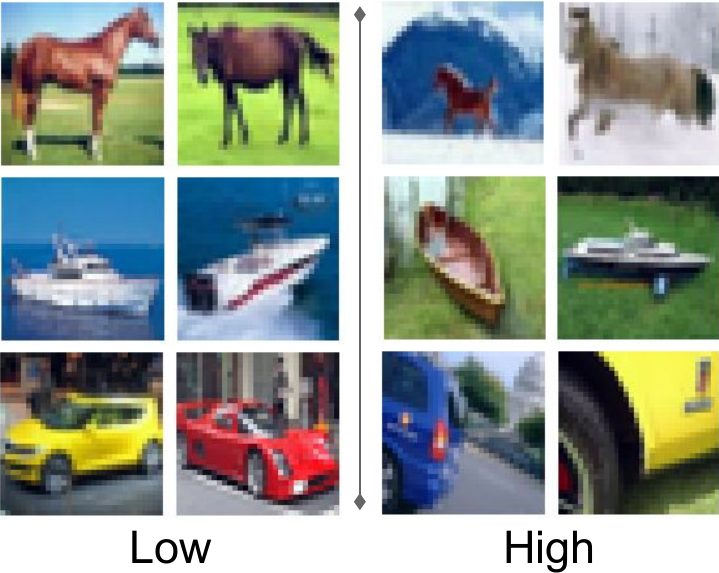}
      {\tiny \CascadedTD} 
    \end{minipage}%
    \begin{minipage}{.35\textwidth}
      \centering
      \includegraphics[width=.94\linewidth]{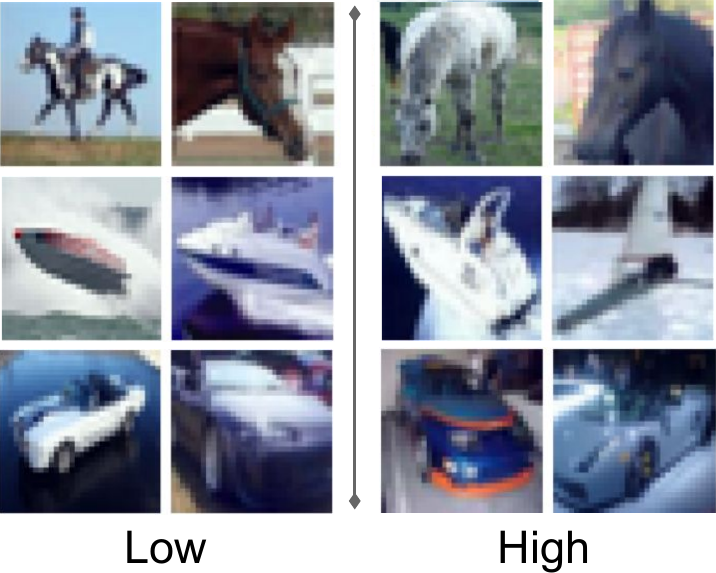}
        {\tiny \CascadedCE} 
    \end{minipage}
    \small
    \captionof{figure}{CIFAR-10 examples of low and high selection latency resulting from \CascadedTD~(\textit{left}) and \CascadedCE~(\textit{right}).}
    \label{fig:cifar_10__qualitative_analysis}
\end{figure}

Figure~\ref{fig:cifar_10__qualitative_analysis} shows CIFAR-10 instances with low and high \textit{selection latency} for both \CascadedTD~and \CascadedCE~models. As with CIFAR-100, the qualitative differences between low and high selection latency for \CascadedTD~are stark, with low selection latency instances being more representative of prototypical instances of the given class (e.g., boats on blue water; horses in fields), whereas high selection latency instances are less typical (e.g., boats on green grass; horses in snow). In contrast, the strong delineation between low and high selection latency groups is not observed for \CascadedCE, supporting the claim that TD training allows the model to more rapidly respond to prototypical exemplars.

\subsection{Generalization to High Resolution Images}
\label{appendix:imagenet_results}
\begin{figure*}[tb]
    \centering
    \includegraphics[width=\textwidth]{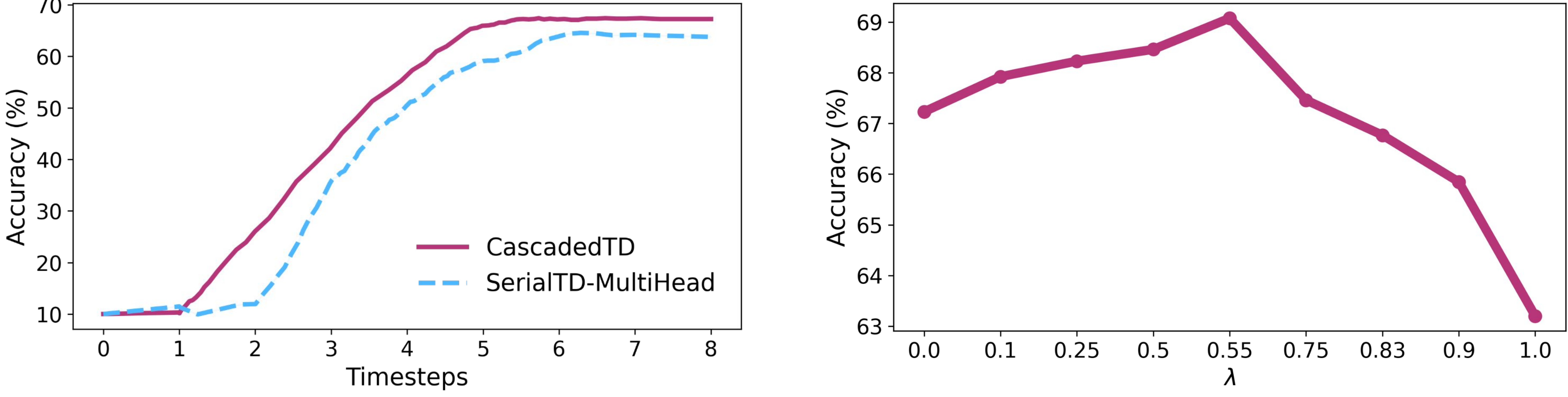}
    \caption{ImageNet2012 Results. Left: Speed accuracy trade off for terrier breed subset of ImageNet2012, obtained by varying a stopping threshold and measuring mean latency and mean accuracy. \CascadedTD\ is our parallel anytime prediction model; \SerialTDMH\ is the state-of-the-art method, SDN \cite{kaya2019shallow}. Right: Effect of TD hyperparameter $\lambda$ on \CascadedTD\ test accuracy. $\lambda = 1$ corresponds to the training methodology of all past research on anytime prediction, which is inferior to any $\lambda < 1$.}
    \label{fig:imagenet_terriers_results}
\end{figure*}

To assess how well our methods generalize to high resolution images, we trained a model on $224\times224$ resolution images of 10 breeds of terriers (dogs) from ImageNet2012 \cite{ILSVRC15}. These breeds are visually similar to one another and cannot be discriminated perfectly based on any simple feature such as color. We observed the same qualitative behavior from models as we observed for our models trained on smaller resolution images (see Figure \ref{fig:imagenet_terriers_results} (left)). We obtain an asymptotic test accuracy of 67.2\% for \CascadedTD(0) versus 63.2\% for {\footnotesize \sc \colorSerialTD{SerialTD(1)-MultiHead}} (i.e., Shallow-Deep Nets), and a 13\% speed up for \CascadedTD(0) to reach a threshold of 50\% accuracy; see Figure \ref{fig:imagenet_terriers_results} (left). 

We conducted a sweep over hyperparameter $\lambda$ to determine its effect on asymptotic accuracy of \CascadedTD. 
Figure \ref{fig:imagenet_terriers_results} (right) shows results for our terrier subset of ImageNet2012. Our results here are consistent with those from our smaller resolution dataset experiments from the main paper, where the $\lambda$ hyperparameter has a systematic effect on performance such that $\lambda < 1$ yields significantly improved performance. For example, TD(0.55) achieves a performance of 69.1\%, which is a 9.3\% improvement over TD(1). This supports our claim that our method generalizes to higher resolution datasets.

For training details, each class consisted of 1300 examples, which were split 90/10 into train and test examples. All simulation details were the same as our previous work, except batch size had to be reduced from 128 to 32 for the larger images, and, the first convolution layer of the ResNet was changed from 3x3 to 7x7 receptive fields.

\subsection{Computing Infrastructure}
\label{appendix:comp_infra}
We used 8x NVIDIA Telsa V100's on Google Cloud Platform (GCP) for training all \CascadedCE~and \CascadedTD~models; a single V100 was used for all evaluations, and to train \MetaCog~models. All models were implemented in PyTorch v1.5.0, using Python 3.7.7 operating on Ubuntu 18.04.

\begin{table}[htpb]
    \caption{Average runtime for training \CascadedCE\ and \CascadedTD\ over CIFAR-10, CIFAR-100, and TinyImageNet.}
    \label{table:average_runtimes}
    \centering
    \footnotesize
    \begin{tabular}{|c|c|c|}
    \hline
    \textbf{Dataset} & \textbf{Model} & \textbf{\begin{tabular}[c]{@{}c@{}}Average Runtime \\ (hours)\end{tabular}} \\ \hline
    \multirow{2}{*}{CIFAR-10} & \CascadedCE & 1.48 $\pm$ 0.002 \\ \cline{2-3} 
     & \CascadedTD & 1.81 $\pm$ 0.001 \\ \hline
    \multirow{2}{*}{CIFAR-100} & \CascadedCE & 1.48 $\pm$ 0.001 \\ \cline{2-3} 
     & \CascadedTD & 1.83 $\pm$ 0.001 \\ \hline
    \multirow{2}{*}{TinyImageNet} & \CascadedCE & 1.45 $\pm$ 0.035 \\ \cline{2-3} 
     & \CascadedTD & 1.97 $\pm$ 0.020 \\ \hline
    \end{tabular}
\end{table}
\subsection{Average Runtime and Reproducibility}
\label{appendix:reproducibility_runtimes}
Table \ref{table:average_runtimes} shows average run times (in hours) for \CascadedCE and \CascadedTD.  Variability in run time, expressed as $\pm1$ SEM, is also shown.
Reproducibility was ensured in the data pipeline and model training by seeding Random, Numpy, and PyTorch packages, as well as flagging deterministic cudnn via PyTorch API. When sweeping over $\lambda$ for a given model and dataset, 5 replications were trained to obtain reliability estimates; a fixed set of 5
seeds was for all models to ensure matched initial conditions across models.
The average runtime for training \MetaCog\ models on a single V100 GPU requires less than 3 minutes. When training multiple trials for a given \MetaCog model, all models are initialized with the same weights, and $42$ was used to seed all packages as detailed above.

\section{Meta-cognitive Experiment Details}
\label{appendix:subsec:metacog}
For all meta-cognitive experiments, training data is generated from the EWS kernel applied to \CascadedTD(0).

\subsection{OOD Detection Dataset Details}
\label{appendix:ood_dataset_details}
CIFAR-10 serves as the in-distribution dataset, which contains 5,000 validation and 10,000 test set instances. The 5,000 validation instances, which we use as the in-distribution training set for OOD, were derived from a 90-10 train-validation split of the original 50,000 training instances used for training the \CascadedTD\ model. The OOD datasets are as follows:

{\textbf{TinyImagenet}} The Tiny ImageNet (TinyImageNet) is a 200-class subset of ImageNet \cite{deng2009imagenet} and it contains 10,000 validation and 10,000 test instances. Following the methods of \citep{liang2017enhancing} we introduce two variations: 1) \textit{resize}; the image is downsampled to $32 \times 32$, and 2) \textit{crop}; a random $32 \times 32$ crop is taken from the image.

{\textbf{LSUN}} The Large-scale Scene UNderstanding (LSUN) \cite{yu2015lsun} consists of 10 scenes categories,  such as classroom, restaurant, bedroom, etc. It contains 10,000 validation and 10,000 test instances, and similar to TinyImageNet, we use the \textit{resize} and \textit{crop} variations.

{\textbf{SVHN}} The Street View House Numbers (SVHN) \cite{netzer2011reading} dataset is obtained from house numbers in Google Street View images. It consists of 73,257 validation and 26,032 test set images.

\subsection{OOD Detection Training Details}
\label{appendix:ood_detection}
The \MetaCog~model is trained for 300 epochs with batch sizes of 256. We used Adam with an initial learning rate of 0.001 and weight decay of 0.0005 to optimize a binary cross entropy loss. Dropout with keep probability 0.5 was used for regularization. Numerical values corresponding to Figure \ref{fig:metacog_ood} are tabulated in Table \ref{table:metacog_ood} with reported SEM corrected to remove random variance \cite{masson2003using}.

The OOD examples from TinyImageNet and LSUN have crop and resize variations \cite{liang2017enhancing} to
make them match CIFAR10 images in dimensions.
\MetaCog\ is trained per (in-, out-of-distribution) dataset pairing---e.g., (CIFAR-10, SVHN)---and input representation type (discussed below). The respective test set is used for evaluation. 

{
\setlength{\tabcolsep}{0.5em}
\begin{table}[htb!]
    \centering
    \small
    \caption{CIFAR-10 (in-distribution) vs. Aggregate OOD dataset quantitative measures corresponding to Figure \ref{fig:metacog_ood}. Each representation may include all time step outputs, $t_{\text{all}}$, or only the final output, $t_\text{final}$.}
     \label{table:metacog_ood}
    \vskip 0.15in
    \begin{tabular}{|c|c|c|}
        \hline
        \textbf{OOD Representation} & \textbf{AUROC} & \textbf{FPR @ 95\% TPR} \\ \hline
        \CascadedTD ~[MSP] & 89.5 $\pm$ 0.5 & 63.0 $\pm$ 3.5 \\ \hline
        \MetaCog ~$t_\text{final}$ [MSP] & 88.8 $\pm$ 0.5 & 63.0 $\pm$ 3.1 \\ \hline
        \MetaCog ~$t_\text{all}$ [MSP] & 90.2 $\pm$ 0.5 & 46.4 $\pm$ 3.1 \\ \hline
        \MetaCog ~$t_\text{final}$ [Entropy] & 90.5 $\pm$ 0.3 & 51.2 $\pm$ 2.5 \\ \hline
        \MetaCog ~$t_\text{all}$ [Entropy] & 92.7 $\pm$ 0.3 & 38.9 $\pm$ 2.5 \\ \hline
        \MetaCog ~$t_\text{final}$ [Softmax] & 92.6 $\pm$ 0.1 & 31.7 $\pm$ 0.7 \\ \hline
        \MetaCog ~$t_\text{all}$ [Softmax] & 95.7 $\pm$ 0.1 & 20.5 $\pm$ 0.7 \\ \hline
        \MetaCog ~$t_\text{final}$ [Logits] & 96.7 $\pm$ 0.1 & 17.5 $\pm$ 0.4 \\ \hline
        \MetaCog ~$t_\text{all}$ [Logits] & 97.3 $\pm$ 0.1 & 13.7 $\pm$ 0.4 \\ \hline
    \end{tabular}
    \vskip -0.1in
\end{table}
}

\subsection{Response Initiation}
\label{appendix:response_initiation}

We explored a third stopping criterion, in addition to the 
confidence-threshold and temporal-deadline criteria. The third criterion
was based on a meta-cognitive model that observes the output \emph{sequence}
from cascaded-model updates and uses this sequence to determine
when to stop. To handle sequences, this \MetaCog\ model was an RNN,
specifically a GRU, trained with a logistic output unit that produced
a binary stop/don't-stop decision. In contrast to the 
confidence-threshold criterion, which is based solely on the network output 
at step $t$, \MetaCog\ in principle uses steps $0-t$ to make its decision.
It produces a continuous output in [0,1], and by stopping when the output
rises above a threshold, we can map out a speed-accuracy trajectory analogous
to that obtained with the confidence-threshold criterion.

\MetaCog\ is trained for 300 epochs with a batch size of 256. We used Adam with an initial learning rate of 0.0001 and weight decay of 0.0001 to optimize a binary cross entropy loss.
The supervised target is 1.0 at step $t$ if the output with highest probability at $t$ remains unchanged for all subsequent steps, or 0.0 otherwise. Essentially, the model is trained to predict when additional
compute will change its decision.
To obtain a finer granularity on time steps, we trained and 
evaluated \MetaCog\ with the EWS kernel. 

We trained \MetaCog\ to predict when to stop for both \CascadedCE\ and
\CascadedTD.
\MetaCog\ is trained on the 4,500 instances of the CIFAR-10 validation set that have been processed by the cascaded model, yielding a training set of dimension $4,500 \times 70 \times 10$, where there are $70$ timesteps and $10$ logit values. We generate our evaluation set from the same method above using the \CascadedCE\ model on the 10,000 instance test set. 

\begin{figure}[htb]
    \centering
    \begin{minipage}{.47\textwidth}
      \centering
      \includegraphics[width=0.98\linewidth]{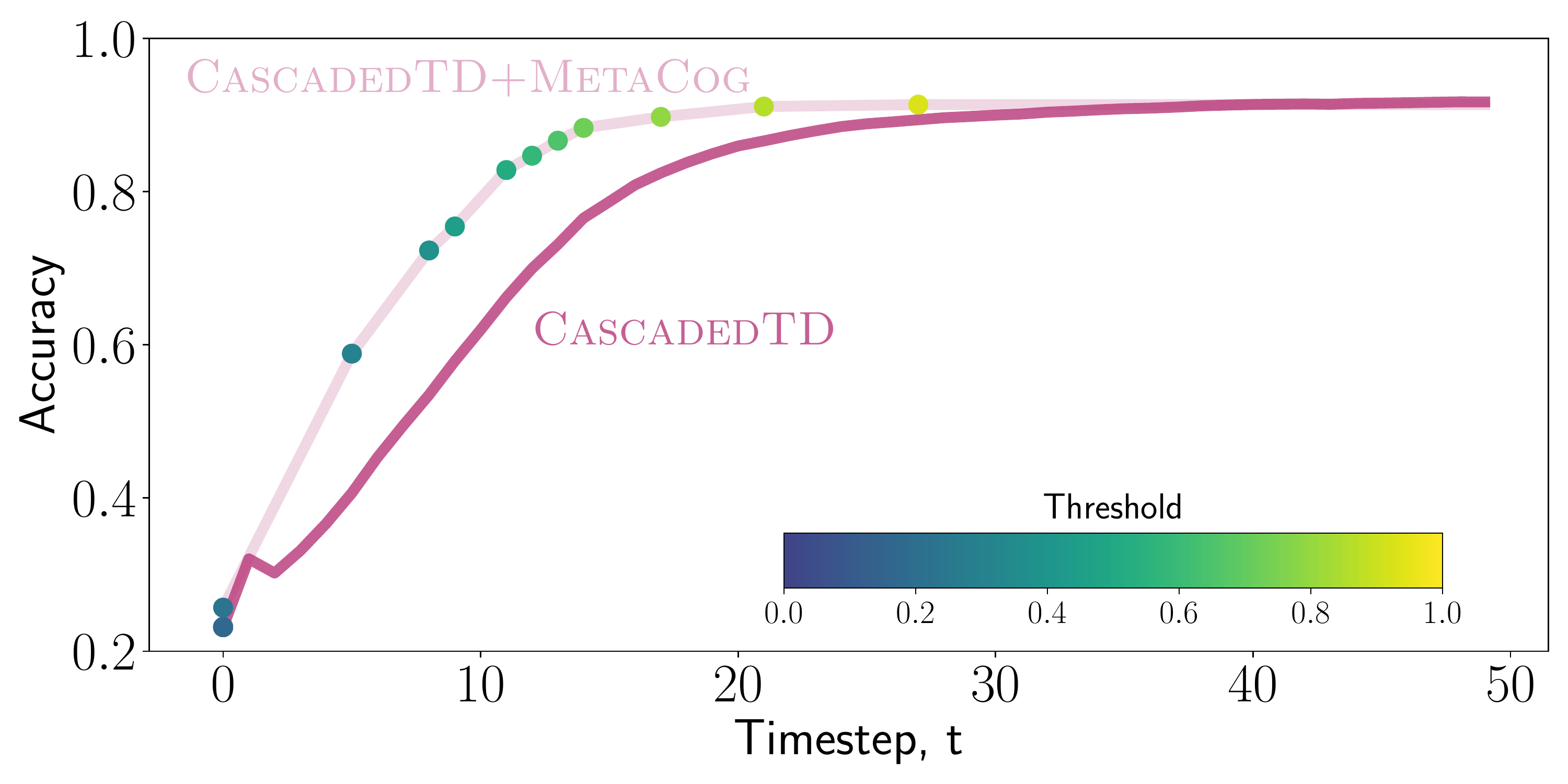}
      {\tiny \CascadedTD} 
    \end{minipage}%
    \begin{minipage}{.47\textwidth}
      \centering
      \includegraphics[width=0.98\linewidth]{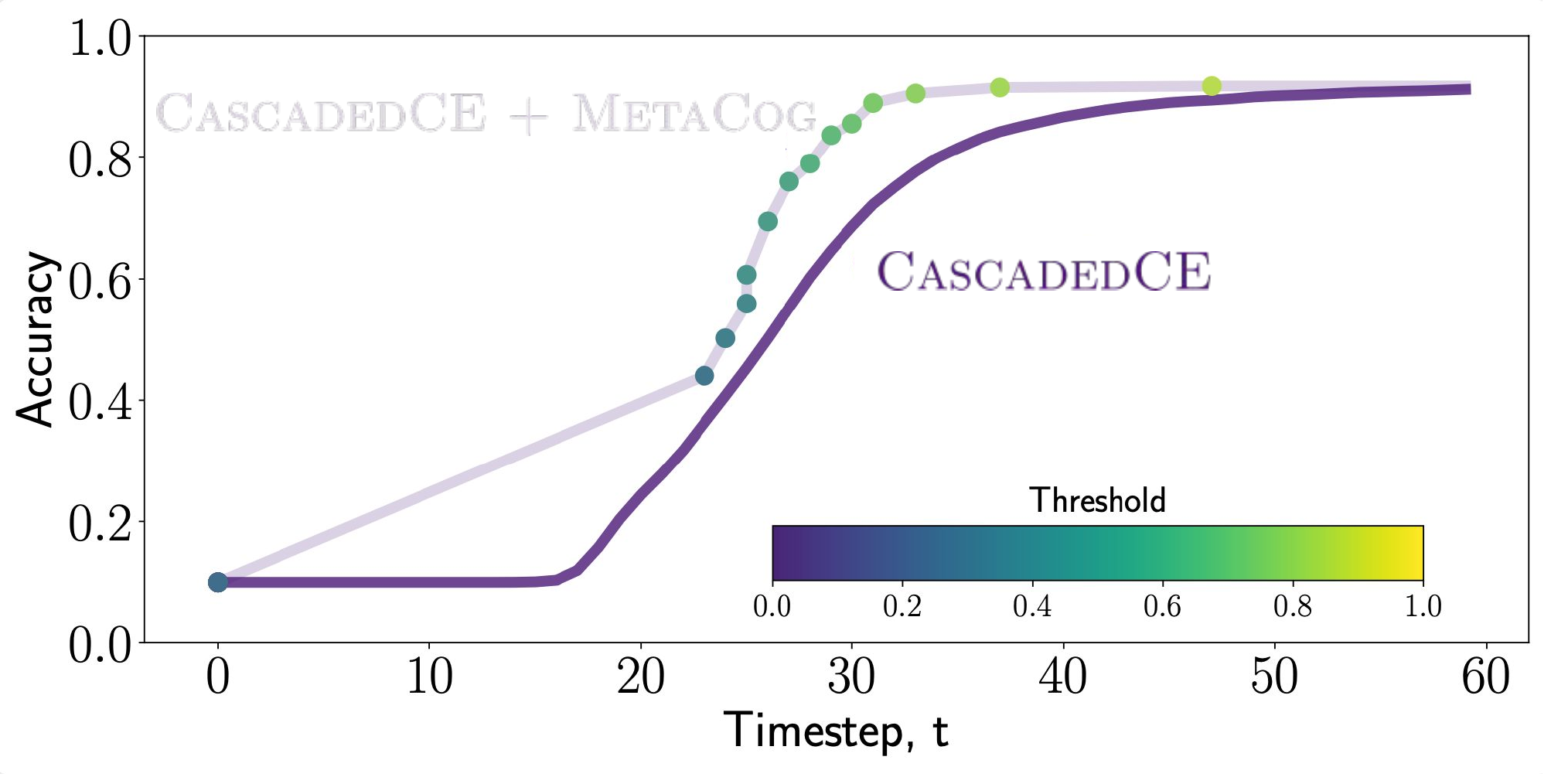}
        {\tiny \CascadedCE} 
    \end{minipage}
    \small
    \captionof{figure}{Response initiation results comparing two stopping
    criteria for \CascadedTD\ (left panel) and \CascadedCE\ (right panel). The solid line represents a temporal-deadline stopping criterion. The fainter dotted line uses \MetaCog~to determine when to stop based on
    an output threshold.}
    \label{fig:rnn__gru__with_labels__CascadedCE}
\end{figure}

Figure \ref{fig:rnn__gru__with_labels__CascadedCE} shows the response initiation results comparing \CascadedTD\ (left panel) and 
\CascadedCE\ (right panel) with stopping criterion using \MetaCog\ versus a temporal deadline. The \MetaCog\ criterion 
yields significant improvements to response initiation for both models.
This finding lends support to the notion that there is a signal in the
model output over time as information trickles through the cascaded layers.
Essentially, \MetaCog\ can interpret the temporal evolution of cascaded
model outputs to improve its speed-accuracy trade off.


\section{Correspondence Between Model Time Steps and Run Time on Parallel Hardware}
\label{appendix:compcomplexity}
The premise of our work is that we have massively parallel hardware with delays on inter-component communication. This architecture allows all ResNet blocks to be updated simultaneously. A block update involves a sequence of matrix multiplications, vector additions, and vector thresholding operations. Because all model variants perform block updates, we needn’t break down the block update into its primitive operations; instead, we consider the basic cycle to be a block update. Due to the assumed communication delay in our hardware, the updated output from a block is not available to other blocks until the next cycle. Both the Serial and Cascaded models can be run on this parallel hardware. The Serial model does not take full advantage of the parallel hardware because it updates only block $k$ at cycle $k$. The Serial model performs anytime read out at cycle $k$ by summing the outputs of blocks 1 through $k-1$ and passing the sum through the classifier head. The Cascaded model takes full advantage of the parallel hardware by updating all blocks at each cycle $k$ using their output from cycle $k-1$. Just as the Serial model, the Cascaded model can perform any-time read out using the one-cycle lagged block outputs. Thus, comparing the Serial and Cascaded models in units of block updates---as we have done throughout the article---provides a runtime comparison on the assumed parallel hardware. Note that Fischer et al. \cite{fischer2018streaming} similarly compare models with different rollouts using the same notion of runtime. 
It is not obvious that parallel updating using partially propagated states will provide benefits.

\section{Checklist}


\begin{enumerate}

\item For all authors...
\begin{enumerate}
  \item Do the main claims made in the abstract and introduction accurately reflect the paper's contributions and scope?
    \answerYes{}
  \item Did you describe the limitations of your work?
    \answerYes{In results and discussion sections}
  \item Did you discuss any potential negative societal impacts of your work?
    \answerNo{We cannot conceive of negative impacts beyond the generic concerns about ML/AI models.}
  \item Have you read the ethics review guidelines and ensured that your paper conforms to them?
    \answerYes{}
\end{enumerate}

\item If you are including theoretical results...
\begin{enumerate}
  \item Did you state the full set of assumptions of all theoretical results?
    \answerYes{}
	\item Did you include complete proofs of all theoretical results?
    \answerYes{Derivations of the fairly trivial theory underlying our model is presented in the supplementary materials.}
\end{enumerate}

\item If you ran experiments...
\begin{enumerate}
  \item Did you include the code, data, and instructions needed to reproduce the main experimental results (either in the supplemental material or as a URL)?
    \answerYes{Details of experimental procedure are in supplemental materials. We will provide a git repository with code once the paper is accepted.}
  \item Did you specify all the training details (e.g., data splits, hyperparameters, how they were chosen)?
    \answerYes{In supplemental materials.}
	\item Did you report error bars (e.g., with respect to the random seed after running experiments multiple times)?
    \answerYes{Where applicable, except for speed-accuracy curves where the error bars were small and the figure was sufficiently complex that error bars would make it harder to interpret figure.}
	\item Did you include the total amount of compute and the type of resources used (e.g., type of GPUs, internal cluster, or cloud provider)?
    \answerNo{We did describe our computing infrastructure in the supplementary materials.}
\end{enumerate}

\item If you are using existing assets (e.g., code, data, models) or curating/releasing new assets...
\begin{enumerate}
  \item If your work uses existing assets, did you cite the creators?
    \answerNA{We do not cite the authors of the three data sets we used, CIFAR-10, CIFAR-100, and TinyImageNet}
  \item Did you mention the license of the assets?
    \answerNA{}
  \item Did you include any new assets either in the supplemental material or as a URL?
    \answerNo{}
  \item Did you discuss whether and how consent was obtained from people whose data you're using/curating?
    \answerNA{}
  \item Did you discuss whether the data you are using/curating contains personally identifiable information or offensive content?
    \answerNA{}
\end{enumerate}

\item If you used crowdsourcing or conducted research with human subjects...
\begin{enumerate}
  \item Did you include the full text of instructions given to participants and screenshots, if applicable?
    \answerNA{}
  \item Did you describe any potential participant risks, with links to Institutional Review Board (IRB) approvals, if applicable?
    \answerNA{}
  \item Did you include the estimated hourly wage paid to participants and the total amount spent on participant compensation?
    \answerNA{}
\end{enumerate}

\end{enumerate}

\end{document}